\documentclass[journal]{IEEEtran}
\usepackage{amsmath,amsfonts,amssymb}
\usepackage{algorithmic}
\usepackage[colorlinks, linkcolor=red, anchorcolor=blue, citecolor=green]{hyperref}
\usepackage[capitalize]{cleveref}
\usepackage{algorithm}
\usepackage{color, colortbl}
\usepackage{array}
\usepackage[caption=false,font=normalsize,labelfont=sf,textfont=sf]{subfig}
\usepackage{textcomp}
\usepackage{stfloats}
\usepackage{url}
\usepackage{verbatim}
\usepackage{graphicx}
\usepackage{cite}

\usepackage{booktabs}
\definecolor{Gray}{gray}{0.9}

\usepackage{multirow}
\usepackage{pifont}

\newcommand{\etal}{\textit{et al.}}
\newcommand{\nbf}[1]{\noindent \textbf{#1.}}
\newcommand{\nit}[1]{\noindent \textit{#1.}}
\newcommand{\pexp}{\noindent where }

\newcommand{\first}[1]{\textcolor{red}{\textbf{#1}}}
\newcommand{\second}[1]{\textcolor{blue}{\textbf{#1}}}

\crefname{section}{Sec.}{Sec.}

\begin{document}
	
\bstctlcite{IEEEexample:BSTcontrol} 

\title{Spectrum-oriented Point-supervised \\ Saliency Detector for Hyperspectral Images}

\author{Peifu Liu, Tingfa Xu$^{\dagger}$, Guokai Shi, Jingxuan Xu, Huan Chen, Jianan Li$^{\dagger}$
    \thanks{Peifu Liu, Tingfa Xu, Jingxuan Xu, Huan Chen, and Jianan Li are with Beijing Institute of Technology, Beijing 100081, China, and with the Key Laboratory of Photoelectronic Imaging Technology and System, Ministry of Education of China, Beijing 100081, China. Email: \{3120245389, ciom\_xtf1, 3120245414, 3220235096, lijianan\}@bit.edu.cn.} %
    \thanks{Tingfa Xu is also with the Big Data and Artificial Intelligence Laboratory, Beijing Institute of Technology Chongqing Innovation Center, Chongqing 401151, China.
    } %
    \thanks{Guokai Shi is with North Automatic Control Technology Institute; Taiyuan 030006, China. Email: guokai\_shi@163.com.} 
    \thanks{$^{\dagger}$ Correspondence to: Tingfa Xu and Jianan Li.}
}%

\markboth{Journal of \LaTeX\ Class Files,~Vol.~14, No.~8, August~2021}%
{Shell \MakeLowercase{\textit{et al.}}: A Sample Article Using IEEEtran.cls for IEEE Journals}


\maketitle

\begin{abstract}
    Hyperspectral salient object detection (HSOD) aims to extract targets or regions with significantly different spectra from hyperspectral images. While existing deep learning-based methods can achieve good detection results, they generally necessitate pixel-level annotations, which are notably challenging to acquire for hyperspectral images. To address this issue, we introduce point supervision into HSOD, and incorporate Spectral Saliency, derived from conventional HSOD methods, as a pivotal spectral representation within the framework. This integration leads to the development of a novel Spectrum-oriented Point-supervised Saliency Detector (SPSD). Specifically, we propose a novel pipeline, specifically designed for HSIs, to generate pseudo-labels, effectively mitigating the performance decline associated with point supervision strategy. Additionally, Spectral Saliency is employed to counteract information loss during model supervision and saliency refinement, thereby maintaining the structural integrity and edge accuracy of the detected objects. Furthermore, we introduce a Spectrum-transformed Spatial Gate to focus more precisely on salient regions while reducing feature redundancy. We have carried out comprehensive experiments on both HSOD-BIT and HS-SOD datasets to validate the efficacy of our proposed method, using mean absolute error (MAE), E-measure, F-measure, Area Under Curve, and Cross Correlation as evaluation metrics. For instance, on the HSOD-BIT dataset, our SPSD achieves a MAE of 0.031 and an F-measure of 0.878. Thorough ablation studies have substantiated the effectiveness of each individual module and provided insights into the model's working mechanism. Further evaluations on RGB-thermal salient object detection datasets highlight the versatility of our approach.  
\end{abstract}


\section{Introduction}

\begin{figure}
    \centering
    \includegraphics[width=\linewidth]{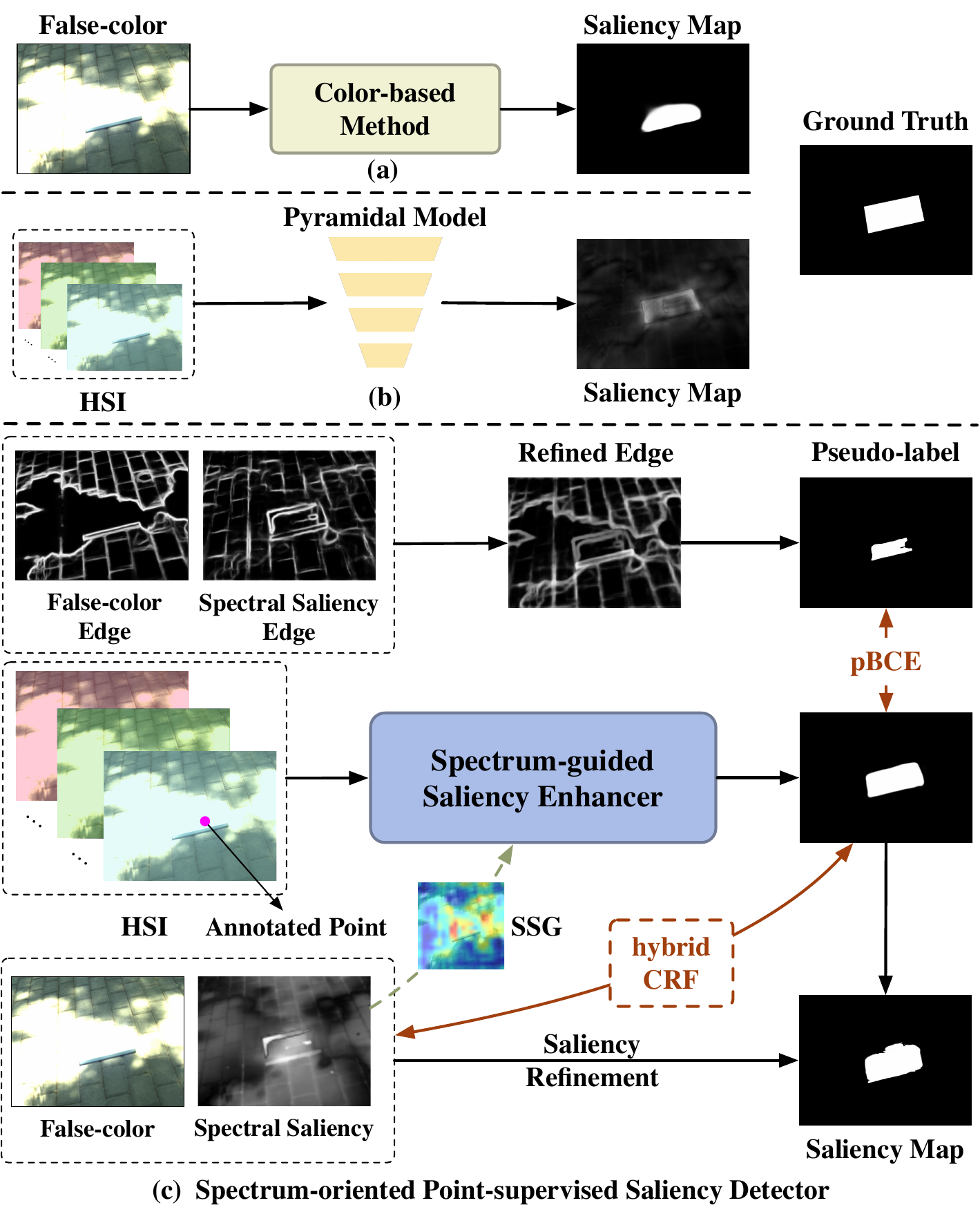}
    \caption{
        HSOD method comparison: (a) Color-based methods struggle under complex lighting; (b) Traditional spectral pyramid methods are hindered by limited feature extraction, resulting in subpar detection; (c) Our Spectrum-oriented Point-supervised Saliency Detector utilizes Spectral Saliency as a foundational element, aiding in the generation of edge-refined pseudo-labels, model supervision, and saliency refinement. The Spectrum-transformed Spatial Gate (SSG) is employed to guide our proposed Spectrum-guided Saliency Enhancer model, enhancing the representation of saliency features.
    }
    \label{fig: fig1}
\end{figure}

\IEEEPARstart{S}{alient} object detection (SOD) aims to develop instruments or systems that mimic the human visual system in identifying and locating the most prominent regions in an image~\cite{10313066, 9371722, HSOD-BIT}. Traditional SOD methods~\cite{itti_1998_a, liu2010learning, 6247743} mainly utilize the inherent color characteristics of RGB imagery to identify salient areas. However, these methods encounter challenges in specific contexts. Distinguishing the foreground from the background can be problematic when they have similar color properties. Additionally, these color-based methods are unable to fully detect targets in cases of overexposure and uneven lighting, as depicted in Fig.~\ref{fig: fig1}~(a).

In contrast to RGB imagery's limitations, hyperspectral images (HSIs) capture a wide array of spatial textures and spectral features simultaneously, making them valuable across various fields~\cite{10261266, 10604291, 9437226, 10613611, LONE2022103752, DSTC}. By analyzing material composition, HSIs can differentiate between foreground and background elements, thereby overcoming the dependency on RGB color attributes alone~\cite{10313066}. This advancement significantly improves saliency detection, particularly in the challenging scenarios previously discussed. The emerging field of hyperspectral salient object detection (HSOD) thus shows considerable promise in various applications, such as target detection~\cite{LONE2022103752}, military surveillance~\cite{Shimoni_2019_military}, and precision agriculture~\cite{SINGH_2020_agriculture}.

Early HSOD methods~\cite{itti_1998_a, liang_2013_salient} focus on ``center-surround" comparisons, accomplished by constructing Gaussian pyramids and assessing spectral similarity across different pyramid layers, as shown in Fig.~\ref{fig: fig1}~(b). These methods rely on hand-crafted features, which limit their feature representation capabilities. Despite a recent deep learning-based approach achieving notable results~\cite{10313066}, it is hindered by its reliance on time-consuming pixel-wise annotations. While numerous studies have investigated weakly-supervised salient object detection (WSOD) techniques for RGB images~\cite{wang_2017_learning, zhang_2020_weaklysupervised, liu_2021_weaklysupervised, gao_2022_weaklysupervised}, the exploration of WSOD methods for hyperspectral images remains an open area of research.

Several low-cost annotation methods are currently available, including image-level annotations~\cite{wang_2017_learning}, scribbles~\cite{zhang_2020_weaklysupervised, 10288361}, bounding boxes~\cite{liu_2021_weaklysupervised}, and points~\cite{gao_2022_weaklysupervised}. Given the lack of a user-friendly annotation tool tailored for HSIs, akin to LabelImg for RGB images, annotation in HSI remains a significant challenge. Consequently, we adopt point annotations as the most rapid and straightforward method, which involves placing one point on the target object and another on the background. 

However, direct application of existing point supervision frameworks~\cite{10.1145/3503161.3547912, gao_2022_weaklysupervised} in hyperspectral images may encounter three issues: \textbf{(i)} These frameworks are not specifically tailored for hyperspectral images, failing to fully exploit the spectrum; \textbf{(ii)} These frameworks utilize the color and structural information of RGB images for model training supervision and saliency refinement. However, the loss of information under certain lighting conditions leads to suboptimal supervision and correction outcomes; \textbf{(iii)} Point supervision is often paired with flood-filling algorithms for pseudo-label generation. The significant edge loss in false-color images can lead the flood-filling algorithm to incorrectly fill the entire image. Constraining the algorithm's range addresses this issue but restricts pseudo-label coverage, ultimately impairing detection performance.

Bearing the above considerations in mind, we conduct an in-depth analysis of Spectral Saliency generated using traditional HSOD methods and discover that under complex lighting conditions, it can effectively maintain the integrity of object. This significant finding prompts us to integrate Spectral Saliency into the point supervision SOD framework and propose an innovative Spectrum-oriented Point-supervised Saliency Detector (SPSD), as illustrated in \cref{fig: fig1} (c). 

\textbf{Firstly}, we propose a novel pipeline for generating pseudo-labels specifically designed for HSIs. By employing Spectral Saliency to refine the edges of the false-color image, we preserved edge integrity. This technique enabled us to broaden the scope of the flood-filling algorithm, consequently expanding the coverage of pseudo-labels and markedly diminishing the adverse effects of point supervision on detection performance. \textbf{Secondly}, we incorporate Spectral Saliency into loss functions and saliency refinement to address the information loss in false-color images. This integration allows the model to more effectively capture structural and edge data during the training. \textbf{Thirdly}, we introduce a Spectrum-transformed Spatial Gate mechanism to improve the model's proficiency in representing salient features. This enhancement sharpens the model's focus on salient areas while minimizing attention to background regions, reducing redundancy in extracted feature, and thus elevating detection accuracy. \textbf{Finally}, we present the Spectrum-guided Saliency Enhancer (SGSE), an innovative solution for efficient and effective hyperspectral salient object detection.

Our methodology is rigorously evaluated using HS-SOD~\cite{imamoglu_2018_hyperspectral} and HSOD-BIT~\cite{HSOD-BIT} datasets. The empirical results clearly indicates that our approach outperforms current leading HSOD methods. Notably, it excels in scenarios marked by similar color, uneven lighting, and overexposure. Furthermore, our method demonstrates impressive effectiveness in RGB-thermal SOD task, highlighting its broad applicability. These results underscore the efficacy and adaptability of our proposed methodology. In a nutshell, our contributions can be summarized as follows:

\begin{itemize}
    \setlength{\itemsep}{0pt}
    \setlength{\parsep}{0pt}
    \setlength{\parskip}{0pt} 
    \item We introduce the innovative Spectrum-oriented Point-supervised Saliency Detector. To the best of our knowledge, it marks the first foray into point supervised hyperspectral salient object detection methods.
    \item We integrate spectrum into generating pseudo-label, making it more apt for hyperspectral images and significantly improving HSOD performance.
    \item We present a novel Spectrum-transformed Spatial Gate mechanism, effectively promoting the model's capability of saliency feature representation.
    
\end{itemize}

\section{Related Work}
\label{sec: related-work}

\subsection{Salient Object Detection}
Traditional SOD methods predominantly rely on hand-crafted features to generate saliency maps~\cite{itti_1998_a, liu2010learning, 6247743}. For example, Itti~\etal~\cite{itti_1998_a} amalgamated color, intensity, and orientation differences across multiple scales to create saliency maps. Machine learning methods have been developed to couple hand-crafted features with techniques such as random forest~\cite{Jiang_2013_CVPR}, boosted decision tree~\cite{Kim_2014_CVPR}, manifold ranking~\cite{6619251}, and minimum spanning tree~\cite{Tu_2016_CVPR}, among others. While these methods can facilitate real-time detection, they often encounter constraints in feature representation, resulting in less than optimal performance and imprecise saliency detection. Recent progress in SOD involves deep learning, particularly CNN integration~\cite{Lee_2016_CVPR, 8315520, li_2017_cnn, kim_2016_a, zhang_2022_r2net, 9756227}. Li \etal~\cite{li_2017_cnn} improved the R-CNN model with local response normalization, enhancing generalization and learning. Kim and Pavlovic~\cite{kim_2016_a} developed a dual-branch CNN for capturing coarse and fine image patch representations. Zhang \etal~\cite{zhang_2022_r2net} utilized a residual pyramid network with dilated convolutions for detailed semantic and feature information. Additionally, attention mechanisms have significantly enhanced SOD performance~\cite{zhang_2018_progressive, liu_2021_visual, 10179145, liu_2022_swinnet}. Zhang \etal~\cite{zhang_2018_progressive} applied spatial and channel-wise attention for contextual information integration, while Liu \etal~\cite{liu_2021_visual} introduced the Visual Saliency Transformer, employing transformers for global context propagation. 

Recently, SOD has expanded to RGB-depth~\cite{liu_2022_swinnet, 10288361}, RGB-thermal~\cite{Tu_2022_RGBT, 9803225}, and video data~\cite{10210391, 10093043}. Despite these advancements, most methods are optimized for RGB images,  with limited effectiveness on HSIs. Our approach fills this gap, offering an HSI-optimized solution.

\subsection{Hyperspectral Salient Object Detection}
Conventional HSOD methods primarily uses low-level hand-crafted features~\cite{le2011saliency, liang_2013_salient, liang2018material}. For instance, Liang \etal~\cite{liang_2013_salient} implemented several strategies derived from Itti's visual model~\cite{itti_1998_a}. These included transforming HSIs to RGB for applying Itti's model, replacing the double color component with grouped band components, or using Euclidean or spectral angle distance for spectral signature matching.

Modern HSOD methods combine features extracted by CNN with machine learning methods, effectively improving the detection performance~\cite{gao_2017_matrix, shen_2019_look, imamoglu_2019_salient, huang_2021_salient, 9941145}. Gao \etal~\cite{gao_2017_matrix} introduced matrix decomposition, using spectral gradient features for low-rank background matrix decomposition to adapt to uneven illumination. Shen \etal~\cite{shen_2019_look} applied an auto-encoder for abundance extraction, with the fraction vector norm indicating spectral purity at each pixel, aiding in salient pixel identification. Imamoglu \etal~\cite{imamoglu_2019_salient} utilized manifold ranking on self-supervised CNN features, and Huang \etal~\cite{huang_2021_salient} combined spatial and spectral feature extraction in CNNs with parallel branches. Koushikey~\etal~\cite{9941145} integrate an extended morphological profile with CNN, which allows for simultaneous utilization of information from adjacent pixels and high-level features. The aforementioned models utilize a limited number of convolutional layers, thereby restricting the model's capacity for feature extraction. Additionally, the incorporation of clustering algorithms further constrains the utilization of extracted features. Consequently, they frequently yield unsatisfactory results.

Most recently, Liu \etal~\cite{10313066} developed SMN, the first fully end-to-end trainable deep neural network for HSOD. It processes frequency-dependent components using SSG and SEO and incorporates Mixed-frequency Attention to utilize distinct frequency features. Qin~\etal~\cite{HSOD-BIT} employed knowledge distillation techniques in the context of the HSOD task, developed DMSSN, and introduced the HSOD-BIT dataset. This dataset includes challenges specifically designed for HSOD, such as similar foreground-background colors, uneven illumination, and overexposure. However, these two methods require extensive pixel-level annotations. Our method, by contrast, needs only a single point for annotation. It combines deep neural networks' robust feature extraction capabilities with low-cost point annotation, achieving effective and efficient hyperspectral salient object detection.

\subsection{Weakly Supervised Salient Object Detection}
Weakly supervised SOD methods \cite{wang_2017_learning, zhang_2020_weaklysupervised, liu_2021_weaklysupervised, piao_2021_mfnet, gao_2022_weaklysupervised, 10288361, 10.1145/3503161.3547912, Yu_Zhang_Xiao_Lim_2021} have attracted significant interest due to their potential to reduce annotation costs using sparse annotations.
Wang \etal~\cite{wang_2017_learning} introduced a two-stage weakly supervised SOD approach employing image-level annotations. Their method utilized a Foreground Inference Network to predict foreground regions, enhancing network performance through fine-tuning.
Zhang \etal~\cite{zhang_2020_weaklysupervised} developed a technique incorporating an auxiliary edge detection branch and scribble annotations to delineate salient objects. This yielded saliency maps with more defined edges, while a gated structure-aware loss improved network structure.
Yu~\etal~\cite{Yu_Zhang_Xiao_Lim_2021} proposed a Saliency Structure Consistency Loss, specifically designed for weakly supervised saliency detection, based on scribble annotations.
Liu \etal~\cite{liu_2021_weaklysupervised} implemented a multi-task map refinement network using bounding boxes, generating initial pseudo labels through an unsupervised approach. These labels were refined and used for training a saliency detector.
Piao \etal~\cite{piao_2021_mfnet} devised a multi-filter directive network. It comprised a directive filter to extract and refine saliency cues from noisy pseudo labels, followed by a saliency detection network to fully leverage these cues.
Despite these methods' effectiveness, the annotation process remains time-intensive.

Recently, Gao \etal~\cite{gao_2022_weaklysupervised} proposed a method using point annotations and a Non-Salient Suppression strategy for network training optimization. However, their constrained flood-filling algorithm led to limited pseudo-label coverage and subpar detection results. Our method redefines the generation of pseudo-labels, achieving extensive coverage without limiting flood-filling range. This significantly reduces detection performance loss typically associated with point supervision.

\subsection{Spectral Saliency}
Spectral Saliency, initially conceptualized by Le Moan \etal~\cite{Le_2013_Saliency}, is defined as a cluster of pixels in an image distinguished primarily by spectral characteristics rather than color attributes. Their model is based on Itti's visual model, which involves calculating center-surround similarities via Gaussian pyramids. These processed maps are subsequently fused and normalized to create a saliency map, epitomizing Spectral Saliency. Zhao \etal~\cite{zhao_2021_salient} further advanced these saliency maps by integrating Spectral Saliency with frequency-tuned saliency detection techniques. This amalgamation yielded saliency maps with more defined boundaries.

Previous research predominantly aligned Spectral Saliency with low-level feature-based methods, often compromising saliency map accuracy. Our approach treats Spectral Saliency as a spectral representation, and incorporates it into the weakly-supervised framework for better hyperspectral image adaptation. The Spectrum-transformed Spatial Gate is also employed to selectively amplifying saliency features.

\section{Method}
\label{sec: method}
Given an HSI denoted as $\boldsymbol{I} \in \mathbb{R}^{\mathrm{H} \times \mathrm{W} \times \mathrm{D}}$, the objective of hyperspectral salient object detection is to generate a corresponding saliency map $\boldsymbol{M} \in \mathbb{R}^{\mathrm{H} \times \mathrm{W} \times \mathrm{1}}$ that highlights the spatial locations of salient objects within the HSI:
\begin{equation}
    \label{eq: mapping func two inputs}
    \boldsymbol{M} = \boldsymbol{\Phi} (\boldsymbol{I}),
\end{equation}
\pexp $\boldsymbol{\Phi}(\cdot)$ signifies the mapping function that ingests HSIs and outputs the corresponding saliency maps. We introduce the Spectrum-oriented Point-supervised Saliency Detector (SPSD) to facilitate this mapping. The entire methodology is divided into three sequential phases: \textbf{(i)} Edge-Refined Pseudo-label Generation; \textbf{(ii)} salient object detection via Spectrum-Guided Saliency Enhancer; and \textbf{(iii)} Saliency Refinement. 

\begin{figure*}[ht]
    \centering
    \includegraphics[width=\linewidth]{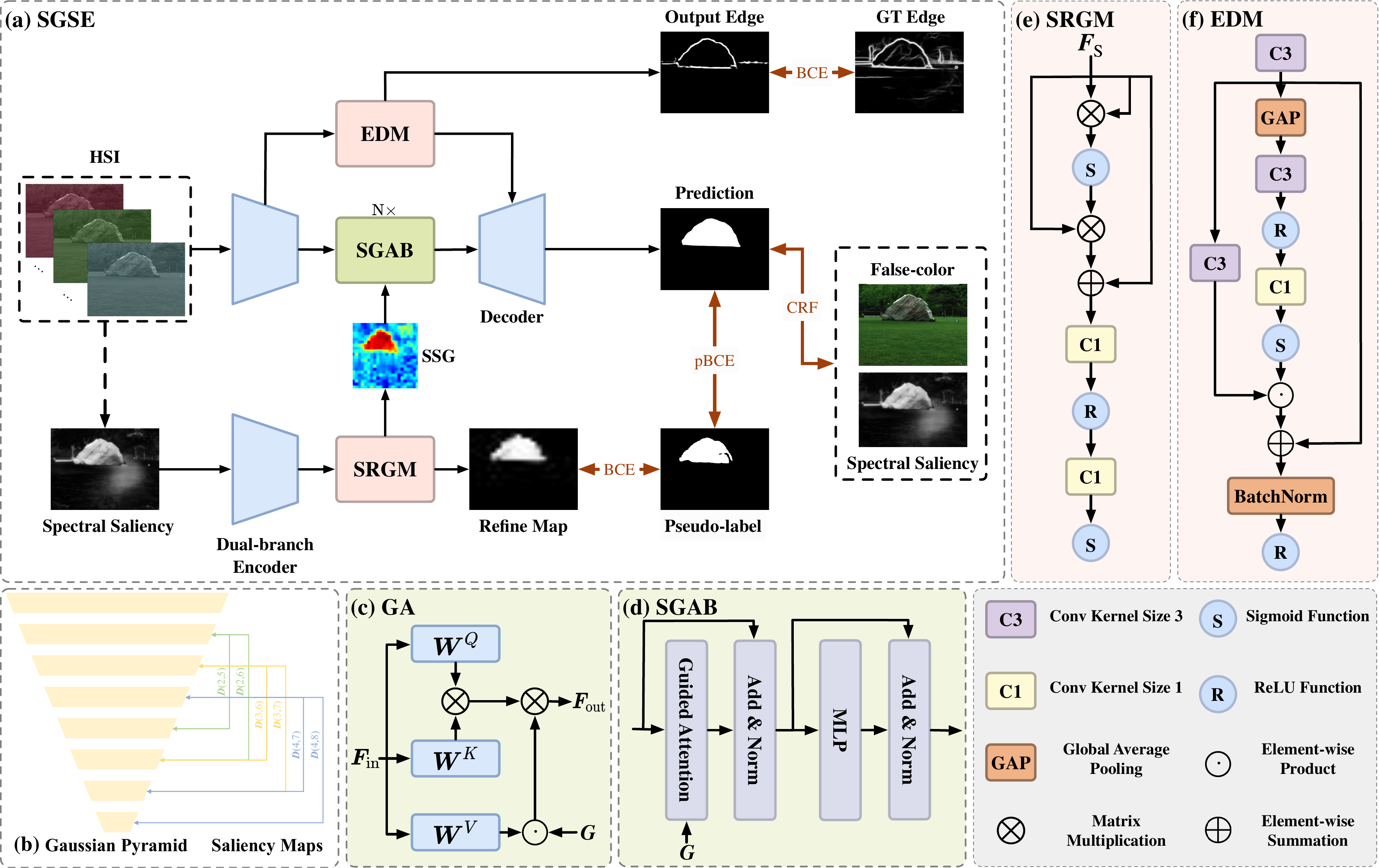}
    \caption{Our proposed (a) Spectrum-guided Saliency Enhancer (SGSE) adheres to an encoder-decoder architecture. Spectral Saliency is obtained by calculating (b) the inter-level differences of the Gaussian pyramid, and subsequently subjected to feature extraction and processed through the proposed (e) Spatial Refinement and Gating Mechanism (SRGM). This mechanism establishes the Spectrum-transformed Spatial Gate (SSG) within the (d) Saliency-Guided Attention Block (SGAB), which leverages (c) Guided Attention (GA) to selectively enhance salient features. Additionally, edge information is preserved via the (f) Edge Detection Module (EDM) and integrated into decoder. The brown arrows in the diagram symbolize supervision.}    
    \label{fig: overall}
\end{figure*}

\subsection{Generating Spectral Saliency Maps}
Compared to traditional saliency maps mainly based on spatial or color information, spectral saliency maps are obtained by constructing Gaussian pyramids and analyze spectral differences at multiple scales~\cite{zhao_2021_salient, liang_2013_salient}, as shown in~\cref{fig: overall}~(b).
Specifically, the input HSI is downsampled through depth-wise convolution using fixed-weight based on Gaussian kernels $\boldsymbol{G}$:
\begin{equation}
    \boldsymbol{G} = \boldsymbol{g}^\intercal \boldsymbol{g},
\end{equation}
where $\boldsymbol{g}=[\frac{1}{16}, \frac{4}{16}, \frac{6}{16}, \frac{4}{16}, \frac{1}{16}]$. Utilizing this kernel, we construct a Gaussian pyramid comprising $\mathrm{K}$ layers (with $\mathrm{K}=9$). Each layer represents a different scale, gradually reducing in spatial dimensions as the scale ascends. The operations of Gaussian blurring and downsampling result in each pixel's information being influenced by an increasingly broader neighborhood of pixels. Subsequently, we compute the spectral angle distance~\cite{liang_2013_salient} between fine and coarse scales to analyze the ``center-surround" dissimilarity, yielding a saliency map $\boldsymbol{D}(c,s)$:
\begin{equation}
    \label{eq: single SS}
    \boldsymbol{D}(c,s)=\text{arccos} \left ( \frac{\boldsymbol{W}_c^\intercal \boldsymbol{W}_s}{\left | \boldsymbol{W}_c \right | \left | \boldsymbol{W}_s \right | } \right ).
\end{equation}
Here, $\boldsymbol{W}_c$ and $\boldsymbol{W}_s$ denote the fine ``center" scale $c$ and coarse ``surround" scale $s$ layers of the pyramid, respectively. Following Itti~\etal~\cite{itti_1998_a}, $c$ is selected from the set $\{ 2,3,4 \}$, and $s$ is determined as either $c+3$ or $c+4$. The final Spectral Saliency map is generated by summing these individual saliency maps:
\begin{equation}
    \label{eq: final SS}
    \boldsymbol{S} = \sum_{c=2}^4 \sum_{s=c+3}^{c=4} \boldsymbol{D}(c,s).
\end{equation}

\subsection{Edge-refined Pseudo-label Generation}
Generating pseudo-labels involves a four-step process: point annotation, edge extraction, false-color edge refinement, and flood filling.

\begin{figure}
    \centering
    \includegraphics[width=\linewidth]{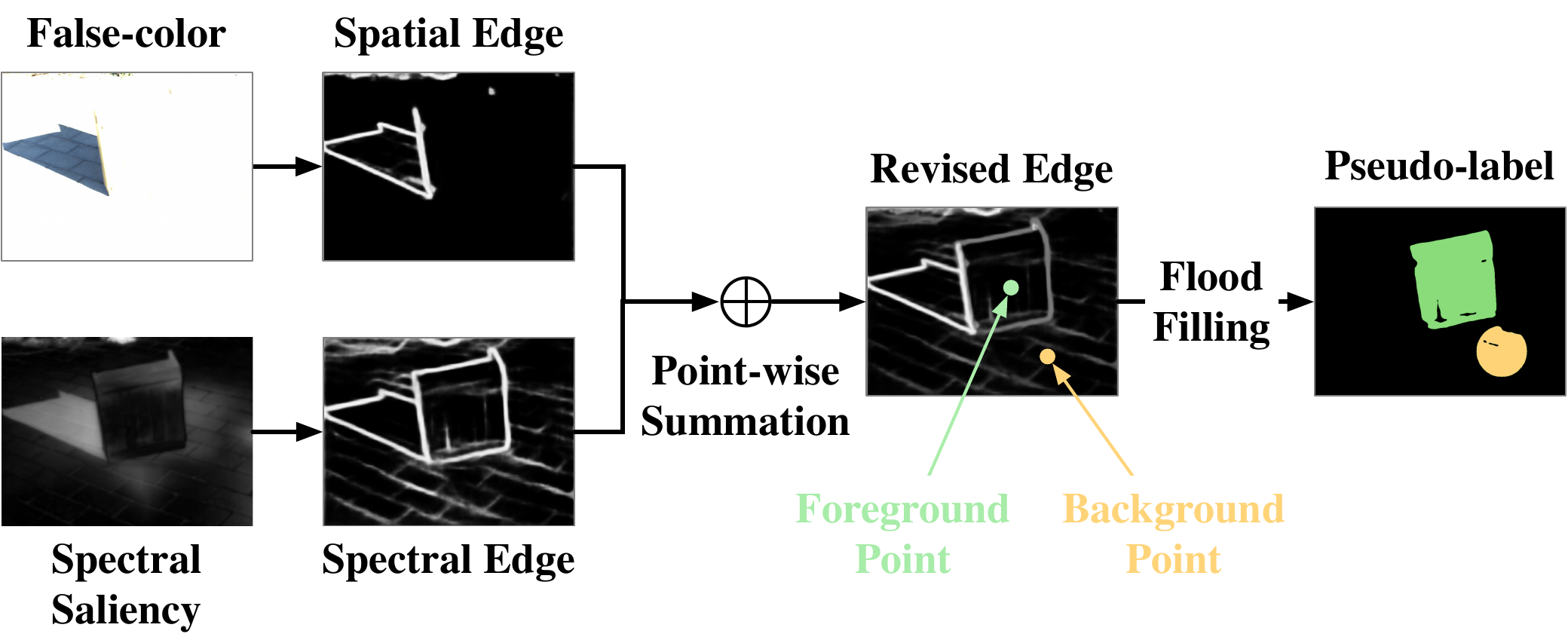}
    \caption{Illustration of Edge-refined Pseudo-label Generation.} 
    \label{fig: gen pseudo-label}
\end{figure}

\nbf{Point Annotation}
Following Gao~\etal~\cite{gao_2022_weaklysupervised}, we annotate one point on each salient target and a point on the background, as depicted in~\cref{fig: gen pseudo-label}. These points are represented as $P=\{ P_\text{S}^i, P_\text{B} | i=1,...,\mathrm{O} \}$, where $P_\text{S}^i$ and $P_\text{B}$ denote the coordinates of the $i$-th salient object (totally $\mathrm{O}$ objects) and the background, respectively. We adopt this annotation strategy because it facilitates the use of the flood fill algorithm for generating pseudo-labels. This algorithm, starting from an initial point, automatically fills the connected region around that point. To ensure accurate identification and segmentation of multiple independent salient objects within an image, it is crucial to manually select a distinct starting point for each target. Without this precaution, the flood fill algorithm might only generate a pseudo-label for a single target, overlooking others. This manual marking process guarantees that each salient target is precisely recognized and segmented, even if they are spatially separate.

\nbf{Edge Extraction}
Following annotation, an edge detector PidNet~\cite{su_2021_pixel} is utilized to extract edge images for flood filling. To improve the delineation of object contours, input images are resized to half their original dimensions prior to edge extraction. This resizing ensures that the generated pseudo-labels encompass maximum coverage of the target regions. Edge images derived from the false-color image and the Spectral Saliency are represented as $\boldsymbol{E}_{I}$ and $\boldsymbol{E}_{S}$, respectively.

\nbf{False-color Edge Refinement}
In cases where edge images from false-color images present discontinuities, using a flood-filling algorithm could lead to the inundation of the entire image. To address this issue, we incorporate Spectral Saliency for edge refinement. The resulting refined edge map, $\boldsymbol{E}_{R}$, is calculated as the sum of the previously extracted edge images:
\begin{equation}
    \label{eq: gen edge}
    \boldsymbol{E}_{R} = \boldsymbol{E}_{I} + \boldsymbol{E}_{S}.
\end{equation}

\nbf{Unrestricted Flood Filling}
The refined edge map $\boldsymbol{E}_{R}$, along with the manually annotated points $P$, are employed to generate the pseudo-label $\boldsymbol{J}$:
\begin{equation}
    \label{eq: flood filling}
    \boldsymbol{J} = \boldsymbol{f} (\boldsymbol{E}_R, P).
\end{equation}
Here, $\boldsymbol{f}(\cdot)$ denotes the flood-filling algorithm. It is important to note that we do not limit the range of the flood-filling process, thereby achieving extensive coverage with pseudo-labels. The complete pseudo-label generation process is depicted in \cref{fig: gen pseudo-label}.

\subsection{Spectrum-guided Saliency Enhancer}
We introduce Spectrum-guided Saliency Enhancer (SGSE) for salient object detection, as depicted in \cref{fig: overall}~(a). 
\textbf{First}, the Spectral Saliency map $\boldsymbol{S}$ and the HSI $\boldsymbol{I}$ are each input into distinct encoder branches for individual feature extraction.
\textbf{Second}, high-level feature derived from the Spectral Saliency map is transformed into a Spectrum-transformed Spatial Gate (SSG) through a Spatial Refinement and Gating Mechanism. The spatial gate modulates the saliency features, concentrating them more significantly within the target areas.
\textbf{Third}, the enhanced saliency features are decoded to obtain the final saliency map $\boldsymbol{M}$.

\nbf{Dual-branch Encoder} 
Our encoder incorporates a dual-branch architecture, comprising two distinct branches: the I-branch, dedicated to the HSI, and the S-branch, focused on the Spectral Saliency map. The deep features extracted from these branches are represented as $\boldsymbol{F}_{I}$ for the I-branch and $\boldsymbol{F}_{S}$ for the S-branch, respectively.

\nit{Backbone Network Encoding}
The mathematical formulation of the S-branch is presented as follows:
\begin{equation}
    \label{eq: S-encoder}
    \boldsymbol{F}_S = \boldsymbol{E}_{\varphi}(\boldsymbol{S}),
\end{equation}
where $\boldsymbol{E}_{\varphi}(\cdot)$ denotes the encoding function parameterized by $\varphi$.

Considering the rich spatial information encapsulated in hyperspectral cubes, we extract an intermediate feature $\boldsymbol{F}_\text{inter}$ from the input HSI. This feature is further processed through an Edge Detection Module (EDM)~\cite{li_2022_multisource} to generate an edge feature. Therefore, the processing within the I-branch is articulated as:
\begin{equation}
    \label{eq: I-encoder}
    \boldsymbol{F}_{I}, \boldsymbol{F}_\text{inter} = \boldsymbol{E}_{\theta}(\boldsymbol{I}).
\end{equation}
Here, $\boldsymbol{F}_{I}$ represents the deep HSI feature, and $\boldsymbol{E}_{\theta}(\cdot)$ signifies the encoder parameterized by $\theta$. In both branches, we utilize a pre-trained backbone network, specifically ResNet50~\cite{He_2016_CVPR}. The backbone architecture remains unmodified in the S-branch. Conversely, in the I-branch, the first convolutional layer's input dimension is modified to fit the HSI input, initializing its weights randomly.

The shapes of $\boldsymbol{F}_{S}$ and $\boldsymbol{F}_{I}$ are identical, denoted as $\mathrm{H}_1 \times \mathrm{W}_1 \times \mathrm{N}$. In this notation, $\mathrm{H}_1 = \frac{\mathrm{H}}{16}$ and $\mathrm{W}_1 = \frac{\mathrm{W}}{16}$ represent the height and width dimensions, respectively, while $\mathrm{N}$ indicates the feature dimension.

\nit{Edge Feature Transformation}  
In higher levels of the feature hierarchy, semantic richness typically increases, often at the cost of finer detail retention~\cite{li_2023_automated}. To retain detailed information, particularly edge details within deeper network layers, we employ an Edge Detection Module to transform the intermediate feature $\boldsymbol{F}_\text{inter}$ into an edge feature $\boldsymbol{F}_{E} \in \mathbb{R}^{\mathrm{4H_1} \times \mathrm{4W_1} \times \mathrm{N'}}$, and generate an output edge map $\boldsymbol{E}_\text{out} \in \mathbb{R}^{\mathrm{H} \times \mathrm{W} \times \mathrm{1}}$. Both the edge feature and the output edge map are supervised using a ground-truth edge map $\boldsymbol{E}_{R}$, ensuring the preservation of finer details. Here, $\mathrm{N'}$ denotes the dimension of the edge feature. Detailed components of EDM are illustrated in \cref{fig: overall}~(f).

\nit{Positional Embedding}  
Following the transformation and flattening of feature dimensions, we introduce learnable positional embedding, analogous to those used in the Vision Transformer~\cite{dosovitskiy_2020_an}. The embedding is critical for the ensuing self-attention calculations. Consequently, the dimensions of the deep features $\boldsymbol{F}_{S}$ and $\boldsymbol{F}_{I}$ are altered to $(\mathrm{H}_1 \times \mathrm{W}_1) \times \mathrm{C}$, where $\mathrm{C}$ signifies the embedding dimension.

\nbf{Spatial Refinement and Gating Mechanism}  
While Spectral Saliency is adept at approximating salient target regions, it occasionally suffers from mislabeling or redundancy due to its reliance on hand-crafted features. To overcome these limitations, we have developed a Spatial Refinement and Gating Mechanism (SRGM) that incorporates spatial attention and gating modules.

Specifically, SRGM is comprised of two key components: refinement and gating. As shown in \cref{fig: overall} (e), we utilize a streamlined self-attention mechanism for the refinement process. To preserve features during the rectification stage, a skip connection is introduced:
\begin{equation}
    \label{eq: SRGM-attention}
    \boldsymbol{F}_S^\prime =  \boldsymbol{F}_S + \boldsymbol{f}_a(\boldsymbol{F}_S),
\end{equation}
where $\boldsymbol{f}_a$ denotes a self-attention operation that excludes linear projection. The omission of linear projection significantly reduces computational complexity, thereby enhancing the efficiency of the process. The gating component of the SRGM is formulated as follows:
\begin{equation}
    \label{eq: SRGM-gating}
    \boldsymbol{G} = \boldsymbol{\sigma}(\boldsymbol{f}_{\theta}(\boldsymbol{\xi}(\boldsymbol{f}_{\varphi}(\boldsymbol{F}_{S}^\prime)))),
\end{equation}
in which $\boldsymbol{\sigma} (\cdot)$ and $\boldsymbol{\xi} (\cdot)$ are the Sigmoid and ReLU activation functions, respectively. The functions $\boldsymbol{f}_{\theta}(\cdot)$ and $\boldsymbol{f}_{\varphi}(\cdot)$ represent $1 \times 1$ convolutional layers, parameterized by $\theta$ and $\varphi$, respectively. Following the application of SRGM, the Spectrum-transformed Spatial Gate $\boldsymbol{G}$ demonstrates an improved focus on saliency compared to $\boldsymbol{F}_{S}$. Additionally, a refinement map $\boldsymbol{G}_{\text{ref}}$ is generated. This map is constrained by a pseudo-label $\boldsymbol{J}$, which aids in optimizing the learning of SRGM parameters.

\nbf{Saliency Guided Attention Block}  
We construct Saliency Guided Attention Block (SGAB) based on Guided Attention~\cite{cai_2022_maskguided} , as shown in \cref{fig: overall}~(c), and (d). The Spectrum-transformed Spatial Gate $\boldsymbol{G}$ functions as a mask within the Guided Attention, effectively intensifying the focus on salient regions:
\begin{equation}
    \label{eq: guidance}
    \boldsymbol{V}_\text{r} = \boldsymbol{V} \odot \boldsymbol{G}.
\end{equation}
Here, $\boldsymbol{V}_\text{r}$ represents the reweighted \textit{value} $\boldsymbol{V}$, and $\odot$ is the element-wise product operation. Following this, self-attention, similar to that in ViT~\cite{dosovitskiy_2020_an}, is utilized to selectively enhance the input saliency feature $\boldsymbol{F}_{I}$ of SGAB.

In terms of decoder application, the enriched feature representation is then input into the same decoder as used in PSOD~\cite{gao_2022_weaklysupervised}, which involves progressive spatial scaling. The edge features $\boldsymbol{F}_{E}$ play a crucial role in the construction of a structurally detailed saliency map.

\subsection{Saliency Refinement}
The point supervision strategy results in inadequate preservation of spatial and chromatic structures, thereby affecting the accuracy of the predicted image~\cite{kolesnikov_2016_seed}. To address this issue, post-processing using a dense conditional random field (CRF) is frequently applied~\cite{liu_2021_weaklysupervised, gao_2022_weaklysupervised}. 
Dense CRF employs binary potential functions to describe relationships between pixels. It assigns the same class label to pixels with closely related attributes, such as color values and relative spatial distances, while distinctly different pixels receive varying labels. This process smoothens the result image and achieves finer boundary details.

In the context of HSOD, challenges arise due to issues like overexposure and uneven illumination in false-color images. The direct application of CRF refinement on these images may result in inaccuracies. To overcome this, we incorporate Spectral Saliency in the CRF refinement process. This integration facilitates the extraction of clear boundaries, even under challenging conditions. The final saliency map is computed by intersecting the refined results obtained from both false-color images $\boldsymbol{R}$ and Spectral Saliency $\boldsymbol{S}$.
\begin{equation}
    \label{eq: CRF amendment}
    \boldsymbol{M} = \boldsymbol{f}_{\text{C}}(\boldsymbol{P}, \boldsymbol{R}) \cap \boldsymbol{f}_{\text{C}}(\boldsymbol{P}, \boldsymbol{S}),
\end{equation}
where $\boldsymbol{f}_{\text{C}}(\cdot)$ represents the dense CRF modification process, and $\boldsymbol{P}$ denotes the prediction from our Spectrum-guided Saliency Enhancer model. This modification effectively refines the saliency region's boundaries, ensuring they correspond to the natural segmentation.

\subsection{Learning Objective}
During the training process, supervision is strategically applied to certain intermediate outputs, ensuring the precise extraction of salient features and edge delineation. Concurrently, the final predictions undergo rigorous supervision to maximize their completeness and accuracy. To facilitate this, a composite loss function is utilized, comprising three components: hybrid CRF loss $\mathcal{L}_{\text{hCRF}}$, partial binary cross-entropy loss $\mathcal{L}_{\text{pBCE}}$, and binary cross-entropy loss $\mathcal{L}_{\text{BCE}}$. This combination is formulated as follows:
\begin{equation}
    \label{eq: sum loss}
    \mathcal{L}_{\text{final}}=\mathcal{L}_{\text{hCRF}}+ \mathcal{L}_{\text{pBCE}}+\mathcal{L}_{\text{BCE}}.
\end{equation}

\nbf{Hybrid CRF Loss}
Our approach incorporates a hybrid CRF loss, which combines the CRF losses for both RGB and Spectral Saliency images. Following Yu~\etal~\cite{Yu_Zhang_Xiao_Lim_2021}, the CRF loss is formulated as:
\begin{equation}
    \label{eq: CRF}
    \mathcal{L}_{\text{CRF}}(\boldsymbol{X};\sigma_P,\sigma_I)=\sum_{i} \sum_{j \in K_i} \left | \mathrm{I}_i - \mathrm{I}_j \right | \boldsymbol{F}(i,j;\sigma_P,\sigma_I),
\end{equation}
where $K_i$ represents a $k \times k$ neighborhood around pixel $i$ in the input matrix $\boldsymbol{X}$. The terms $\mathrm{I}_i$ and $\mathrm{I}_j$ denote the saliency values at positions $i$ and $j$, respectively. $\boldsymbol{F}$ is the Gaussian filter defined as:
\begin{equation} \small
    \label{eq: Gaussian filter}
    \boldsymbol{F}(i,j;\sigma_P, \sigma_I) = \frac{1}{w} \text{exp}\left(-\frac{{\left \| P(i) - P(j) \right \|}^2}{2\sigma_P^2}-\frac{{\left \| \mathrm{I}(i) - \mathrm{I}(j) \right \|}^2}{2\sigma_I^2}\right),
\end{equation}
in which $\frac{1}{w}$ is the normalization weight, and $P(\cdot)$ and $\mathrm{I}(\cdot)$ represent the spatial position and RGB value of a pixel in $\boldsymbol{X}$, respectively. The CRF loss is inversely proportional to the spatial proximity and RGB value similarity between two points. The hyperparameters $\sigma_P$ and $\sigma_I$ adjust the Gaussian kernel's influence, balancing the spatial and chromatic information considered during the computation.

The hybrid CRF loss sums the CRF losses for the false-color image $\boldsymbol{R}$ and Spectral Saliency $\boldsymbol{S}$:
\begin{equation}
    \label{eq: hyber CRF}
    \mathcal{L}_{\text{hCRF}}=\mathcal{L}_{\text{CRF}}(\boldsymbol{R};\sigma_P,\sigma_I) + \mathcal{L}_{\text{CRF}}(\boldsymbol{S};\sigma_P^\prime,\sigma_I^\prime),
\end{equation}
with $\sigma_P^\prime$ and $\sigma_I^\prime$ being the hyperparameters for Spectral Saliency.

\nbf{Partial Binary Cross-Entropy Loss} 
The partial binary cross-entropy (pBCE) loss targets definite regions while excluding ambiguous areas:
\begin{equation}
    \label{eq: pBCE}
    \mathcal{L}_{\text{pBCE}}=-\sum_{j \in \boldsymbol{J}}{\left [ \boldsymbol{J}_j \text{log}(\boldsymbol{S}_j)+(1-\boldsymbol{J}_j) \text{log}(1-\boldsymbol{S}_j ) \right ]},
\end{equation}
where $j$ indicates a pixel in the pseudo-label $\boldsymbol{J}$, derived from annotated points and extracted edges. $\boldsymbol{S}$ represents the predicted saliency map.

\nbf{Binary Cross-entropy Loss} 
The binary cross-entropy (BCE) loss is formulated as:
\begin{equation}
    \label{eq: equation BCE}
    \mathcal{L}_{\text{BCE}}(\boldsymbol{X},\boldsymbol{Y})=-\sum {\left [ \boldsymbol{X} \text{log}(\boldsymbol{Y})+(1-\boldsymbol{X}) \text{log}(1-\boldsymbol{Y}) \right ]},
\end{equation}
\pexp $\boldsymbol{X}$ and $\boldsymbol{Y}$ represent the ground truth and input matrix, respectively. BCE constrains two outputs: the output edge map $\boldsymbol{E}_{\text{out}}$ and the refinement map $\boldsymbol{G}_{\text{ref}}$:
\begin{equation}
    \label{eq: BCE application}
    \mathcal{L}_{\text{BCE}} = \mathcal{L}_{\text{BCE}}(\boldsymbol{E}_R,\boldsymbol{E}_{\text{out}})+\mathcal{L}_{\text{BCE}}(\boldsymbol{J},\boldsymbol{G}_{\text{ref}}),
\end{equation}
\pexp $\boldsymbol{E}_R$ and $\boldsymbol{J}$ are the ground-truth edge map and pseudo-label, respectively.

\section{Experiments}
\subsection{Experimental Settings}
\nbf{Datasets}
We evaluate the performance of our SPSD on two datasets: HS-SOD~\cite{imamoglu_2018_hyperspectral} and HSOD-BIT~\cite{HSOD-BIT}. HS-SOD dataset consists of 60 hyperspectral images with a spatial resolution of $768\times1024$ pixels and a spectral resolution of 5 $\rm{nm}$. These images cover a spectral range from 380 to 780 $\rm{nm}$. We manually select 48 images for training, and the rest 12 images for testing.

HSOD-BIT dataset comprises 319 HSIs. Each HSI has a spatial resolution of $1240\times1680$ pixels and spans a spectral range of 400-1000 $\rm{nm}$, with a spectral interval of 3 $\rm{nm}$. We select 64 images from different scenarios as test data. Both datasets are accompanied by corresponding false-color images and binarized ground-truth images, facilitating the evaluation and comparison of saliency detection performance.

\nbf{Evaluation Metrics}
We employ five metrics to gauge the performance of our SPSD: Mean Absolute Error (MAE), E-measure ($E_\xi$), adaptive F-measure ($F_\beta$), Area Under Curve (AUC), and Cross Correlation (CC).

\begin{table*}
    \centering
    \caption{Quantitative results on HSOD-BIT and HS-SOD datasets. Column ``Type" denotes supervision type. FS: fully supervised, WS: weakly supervised, US: unsupervised. Red represents the best, blue represents the second best.}
    \setlength{\tabcolsep}{3.2mm}{
        \begin{tabular}{lccccccccccc}
            \toprule[1.2pt]
            \multicolumn{1}{l|}{\multirow{2}{*}{Method}} & \multicolumn{1}{c|}{\multirow{2}{*}{Type}} & \multicolumn{5}{c|}{HSOD-BIT} & \multicolumn{5}{c}{HS-SOD} \\
            \cline{3-12}
            \multicolumn{1}{c|}{} & \multicolumn{1}{c|}{} & \multicolumn{1}{l}{MAE $\downarrow$} & \multicolumn{1}{l}{$E_\xi$ $\uparrow$} & \multicolumn{1}{l}{$F_\beta$ $\uparrow$} & \multicolumn{1}{l}{AUC $\uparrow$} & \multicolumn{1}{l|}{CC  $\uparrow$}  & \multicolumn{1}{l}{MAE $\downarrow$} & \multicolumn{1}{l}{$E_\xi$ $\uparrow$} & \multicolumn{1}{l}{$F_\beta$ $\uparrow$} & \multicolumn{1}{l}{AUC $\uparrow$} & \multicolumn{1}{l}{CC $\uparrow$} \\ 
            
            \midrule[1.1pt]
            \multicolumn{11}{c}{\textit{RGB-image-based SOD Methods}} \\ 
            \midrule[1.1pt]
            \multicolumn{1}{l|}{EDN-R\cite{wu_2022_edn}} & \multicolumn{1}{c|}{FS} & 0.066 & 0.940  & 0.851 & \first{0.981} & \multicolumn{1}{c|}{\second{0.890}} & 0.236 & 0.729 & 0.566 & 0.942 & 0.625   \\
            \multicolumn{1}{l|}{TRACER~\cite{lee2022TRACER}} & \multicolumn{1}{c|}{FS} & 0.039 & 0.913 & 0.802 & \second{0.970} & \multicolumn{1}{c|}{0.846} & 0.158 & 0.610 & 0.393 & 0.868 & 0.465 \\
            \multicolumn{1}{l|}{BBRF~\cite{Ma2023BBRF}} & \multicolumn{1}{c|}{FS} & \second{0.033} & 0.930 & 0.850 & 0.932 & \multicolumn{1}{c|}{0.845} & 0.090 & 0.753 & 0.538 & 0.781 & 0.520 \\
            \multicolumn{1}{l|}{ABiU\_Net~\cite{Qiu_2023_ABiU_Net}} & \multicolumn{1}{c|}{FS} & 0.037 & \first{0.957} & \second{0.877} & 0.963 & \multicolumn{1}{c|}{\first{0.894}} & 0.119 & 0.620 & 0.391 & 0.846 & 0.472\\
            \multicolumn{1}{l|}{PSOD\cite{gao_2022_weaklysupervised}} & \multicolumn{1}{c|}{WS} & 0.043 & 0.891 & 0.763 & 0.956  & \multicolumn{1}{c|}{0.806} & 0.097 & 0.690 & 0.526 & \second{0.953} & 0.616 \\
            \multicolumn{1}{l|}{PSOD (Our label)} & \multicolumn{1}{c|}{WS} & 0.035 & 0.916 & 0.808 & 0.956  & \multicolumn{1}{c|}{0.835}& \first{0.048} & 0.774 & \first{0.641} & \first{0.983}    & \first{0.731} \\
            \midrule[1.1pt]
            \multicolumn{11}{c}{\textit{HSI-based HSOD Methods}} \\ 
            \midrule[1.1pt]
            \multicolumn{1}{l|}{Itti\cite{itti_1998_a}} & \multicolumn{1}{c|}{US} & 0.249 & 0.732 & 0.343 & 0.801  & \multicolumn{1}{c|}{0.302} & 0.259 & 0.539 & 0.207 & 0.783 & 0.225 \\
            \multicolumn{1}{l|}{SAD\cite{liang_2013_salient}} & \multicolumn{1}{c|}{US} & 0.205 & 0.735 & 0.363 & 0.794 & \multicolumn{1}{c|}{0.264} & 0.205 & 0.546 & 0.197 & 0.778 & 0.223  \\
            \multicolumn{1}{l|}{SED\cite{liang_2013_salient}} & \multicolumn{1}{c|}{US} & 0.131 & 0.753 & 0.345 & 0.794 & \multicolumn{1}{c|}{0.283} & 0.133 & 0.577 & 0.258 & 0.793 & 0.200  \\
            \multicolumn{1}{l|}{SG\cite{liang_2013_salient}} & \multicolumn{1}{c|}{US} & 0.183 & 0.658 & 0.286 & 0.813 & \multicolumn{1}{c|}{0.321} & 0.197 & 0.563 & 0.234 & 0.808 & 0.268  \\
            \multicolumn{1}{l|}{SUDF\cite{imamoglu_2019_salient}} & \multicolumn{1}{c|}{US} & 0.151 & 0.781 & 0.598 & 0.918 & \multicolumn{1}{c|}{0.671} & 0.242 & 0.554 & 0.256 & 0.723 & 0.250 \\ 
            \multicolumn{1}{l|}{SMN-R\cite{10313066}} & \multicolumn{1}{c|}{FS} & 0.039 & 0.922 & 0.827 & 0.969 & \multicolumn{1}{c|}{0.849} & \second{0.070} & \second{0.800} & \second{0.638} & 0.903 & \second{0.684} \\
            \multicolumn{1}{l|}{SPSD (PSOD label)} & \multicolumn{1}{c|}{WS} & 0.041 & 0.885 & 0.812 & 0.895 & \multicolumn{1}{c|}{0.779} & 0.074 & \first{0.802}  & 0.581  & 0.819 & 0.656 \\
            \multicolumn{1}{l|}{SPSD (Ours)} & \multicolumn{1}{c|}{WS} & \first{0.031} & \second{0.944} & \first{0.878} & 0.938 & \multicolumn{1}{c|}{0.864} & 0.073 & 0.778  & 0.626  & 0.846 & 0.630 \\ \bottomrule[1.2pt]
        \end{tabular}
    }
    \label{Dingliang}
\end{table*}

\nbf{Implementation Details}
The model is developed using the PyTorch framework. Optimization is achieved through Stochastic Gradient Descent across 200 epochs, employing a momentum of 0.9 and a weight decay factor of \(5 \times 10^{-4}\), following the hyper-parameter settings used by Gao~\etal~\cite{gao_2022_weaklysupervised}. The peak learning rate is established at \(5\times10^{-3}\), incorporating both warm-up and linear decay strategies for effective learning rate adjustment. In each epoch, data augmentation is applied to each image with a random probability of horizontal flipping, followed by random cropping that does not preserve the original aspect ratio. A consistent batch size of 6 is maintained throughout the training process.

In the context of the hybrid CRF loss, the hyperparameters \(\sigma_I\), \(\sigma_P\), \(\sigma_I^{\prime}\), and \(\sigma_P^{\prime}\) are configured to 0.03, 5, 3, and 0.003, respectively. Input images are uniformly resized to dimensions of \(352 \times 352\) pixels. The dual-branch encoder employs ResNet50~\cite{He_2016_CVPR} as its backbone. Our model integrates four SGABs to enhance intermediate feature representation.

\nbf{Competing methods} 
We benchmark our SPSD against Itti's model~\cite{itti_1998_a} and various established hyperspectral salient object detection methods as proposed by Liang \etal~\cite{liang_2013_salient}. These methods include the spectral angle distance (SAD), spectral Euclidean distance (SED), and spectral grouping (SG). For certain conventional approaches, HSIs are initially converted into RGB images. This process necessitates downsampling the HSIs to 33 channels to align with the dimension of the conversion matrix. In order to facilitate a comprehensive comparison with cutting-edge techniques, our evaluation also encompasses the SUDF~\cite{imamoglu_2019_salient} and SMN~\cite{10313066}. We maintain the default settings of them to ensure an equitable comparison. Experiments involving both traditional methods and SUDF are conducted on the test sets of HSOD-BIT and HS-SOD.

Moreover, our analysis includes fully supervised RGB-image-based SOD methods such as EDN~\cite{wu_2022_edn}, TRACER~\cite{lee2022TRACER}, BBRF~\cite{Ma2023BBRF}, and ABiU\_Net~\cite{Qiu_2023_ABiU_Net}. Additionally, a weakly supervised RGB-image-based method, PSOD~\cite{gao_2022_weaklysupervised}, is examined. To maintain fairness in the comparison, EDN and SMN utilize ResNet as their backbone networks and are therefore referred to as EDN-R and SMN-R, respectively.

To further verify the effectiveness of our edge-refined pseudo-labels, we swapped the pseudo-label generation methods between PSOD and our approach. These two experiments are referred to as PSOD (Our label) and SPSD (PSOD label), respectively.

\begin{figure*}[tp]
    \centering
    \includegraphics[width=\linewidth]{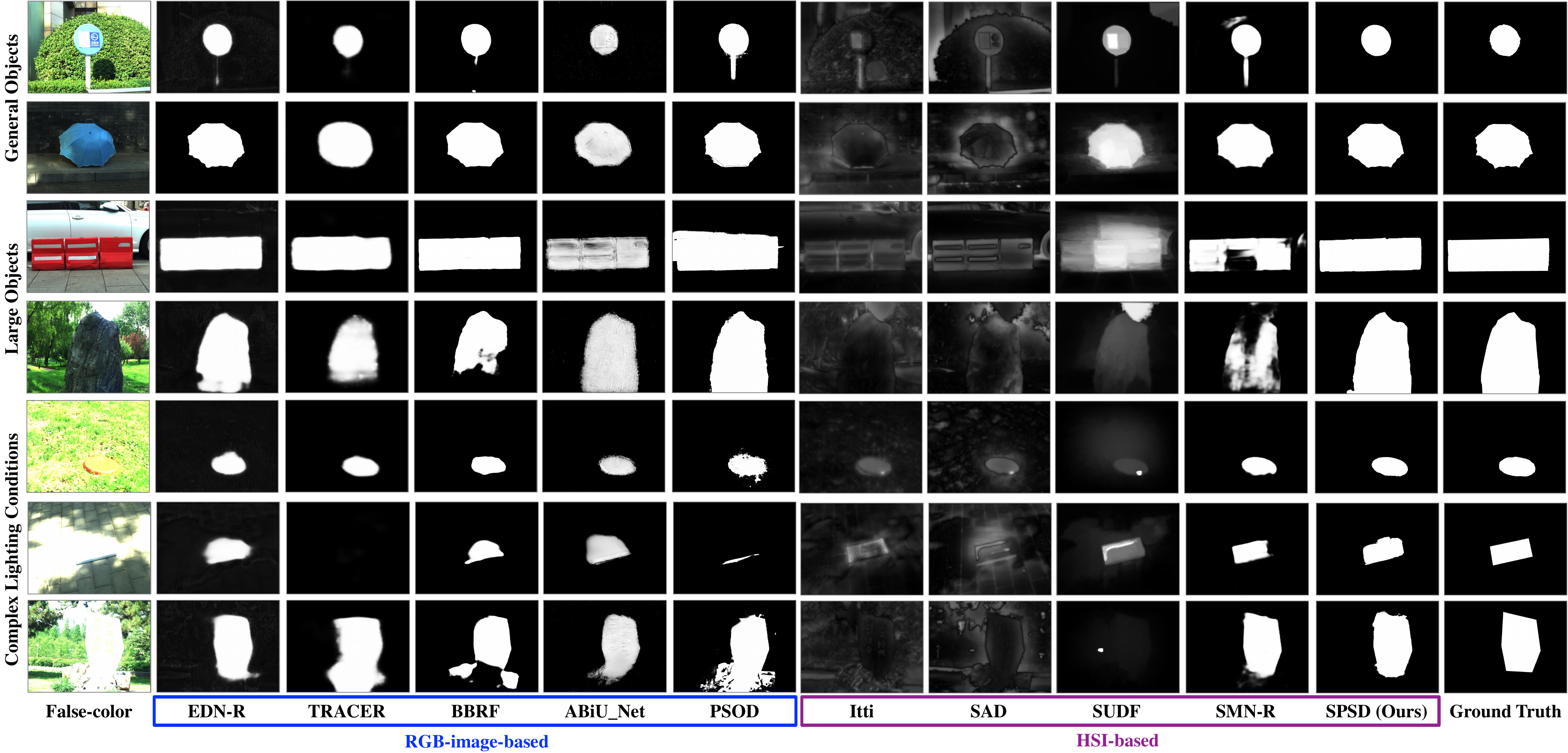}
    \caption{Qualitative results on HSOD-BIT. From top to bottom: general objects, large objects, and objects under complex lighting conditions.}
    \label{fig: dingxing}
\end{figure*}

\subsection{Results on HSOD-BIT}
\nbf{Quantitative Results}
The quantitative analysis on the HSOD-BIT dataset is detailed in \cref{Dingliang}. Our method, SPSD, outperforms both conventional saliency detection techniques and SUDF across evaluation metrics. Specifically, SPSD achieves an $F_\beta$ score of $0.878$ and a CC score of $0.864$, marking a substantial improvement of $46.8\%$ and $28.8\%$ over SUDF, respectively. Traditional methods, which depend on hand-crafted features, are constrained by their limited feature extraction capabilities. Although SUDF incorporates CNNs for feature extraction, its subsequent application of manifold learning and superpixel clustering diminishes the effectiveness of these features, leading to inferior detection performance.

In comparison with SMN-R, SPSD exhibits superior performance, significantly highlighting our method's efficacy in hyperspectral salient object detection. Against fully supervised RGB-image-based methods, SPSD demonstrates competitive results. Specifically, SPSD scores $0.944$ in $E_\xi$, narrowly trailing ABiU\_Net's score of $0.957$, and outperforming other methods. Additionally, it achieves the best results in MAE and $F_\beta$. SPSD outperforms PSOD in all other metrics, except for AUC. This suggests that the transformation of hyperspectral images into pseudo-color images and the subsequent application of existing RGB SOD algorithms is less effective. The superior performance of SPSD underscores the advantages of direct hyperspectral image processing, which allows for more precise saliency detection by fully leveraging spectrum.

Notably, integrating PSOD with the pseudo-labels generated by our method significantly enhances detection performance. For example, the integration reduces MAE by $18.6\%$, and increases $F_\beta$ and CC by $5.77\%$ and $3.60\%$, respectively. Conversely, when utilizing the pseudo-labels generated by PSOD, our method demonstrates a marked decline across all evaluation metrics. These results substantiate the effectiveness of our edge-refined pseudo-label generation process, which incorporates Spectral Saliency to mitigate the shortcomings of point supervision. By enhancing the edges and expanding the coverage of pseudo-labels, SPSD ensures accurate and robust saliency detection.

\begin{table}[tp]
    \centering
    \caption{Quantitative Efficiency Analysis.}
    \label{Efficiency Analysis}
    \setlength{\tabcolsep}{1.8mm}{
        \begin{tabular}{l|cccc} 
            \toprule[1.2pt]
            Method & FLOPs (G) & \#Params (M) & Speed (FPS) & MAE $\downarrow$\\ 
            \midrule
            EDN-R~\cite{wu_2022_edn}  & 20.42    & 33.04        & 33.71       & 0.066 \\
            BBRF~\cite{Ma2023BBRF} & 46.41 & 74.01 & 25.96 & 0.033 \\
            TRACER~\cite{lee2022TRACER} & \textbf{5.20} & 11.09 & 12.27 & 0.039 \\
            ABiU\_Net~\cite{Qiu_2023_ABiU_Net} & 24.89 & 33.50 & 30.79 & 0.037 \\
            PSOD~\cite{gao_2022_weaklysupervised}   & 108.89     & 91.61        & \textbf{40.65} & 0.043 \\ 
            SUDF~\cite{imamoglu_2019_salient}  & 82.90         & \textbf{0.10}            & 0.51          & 0.151 \\ 
            SMN-R~\cite{10313066} & 14.58 & 7.27 & 35.91 & 0.039 \\
            \midrule
            SPSD (Ours)     & 54.68     & 8.09         & 29.00     & \textbf{0.031}  \\
            \bottomrule[1.2pt]
        \end{tabular}
    }
\end{table}

\nbf{Efficiency Analysis}
The computational efficiency of SPSD is evaluated in comparison to other methods, focusing on metrics such as floating point operations (FLOPs), number of parameters (\#Params), and inference speed (FPS). All methods are tested with their spatial and spectral dimensions set to default values.

As detailed in \cref{Efficiency Analysis}, SPSD is a well-balanced model, achieving an optimal compromise between efficiency and effectiveness. Specifically, the FLOPs of SPSD are moderately scaled, suggesting that it is less computationally demanding than some high-performance methods like PSOD and SUDF, which have FLOPs of $108.89$~G and $82.90$~G, respectively. Despite possessing considerably fewer parameters - only $8.09$~M, in contrast to PSOD's $91.61$~M and SUDF's $0.10$~M - SPSD sustains a competitive inference speed of $29.00$~FPS. This rate approaches that of the fastest models. Moreover, SPSD's exceptional accuracy, as evidenced by the lowest MAE of $0.031$, underscores its superior detection capabilities.

\nbf{Qualitative Results}
Qualitative results on the HSOD-BIT dataset are illustrated in \cref{fig: dingxing}. For general objects, SPSD demonstrates performance comparable to the latest state-of-the-art methods in both RGB and HSI domains. It captures the complete extent of the objects and delineates clear boundaries. In scenarios involving large objects, SPSD distinguishes itself from methods like TRACER, BBRF, SUDF, and SMN-R by not omitting significant portions of the object. This superior performance can be attributed to our method's ability to effectively utilize the full spectrum.

\begin{figure*}[htp]
    \centering
    \includegraphics[width=\linewidth]{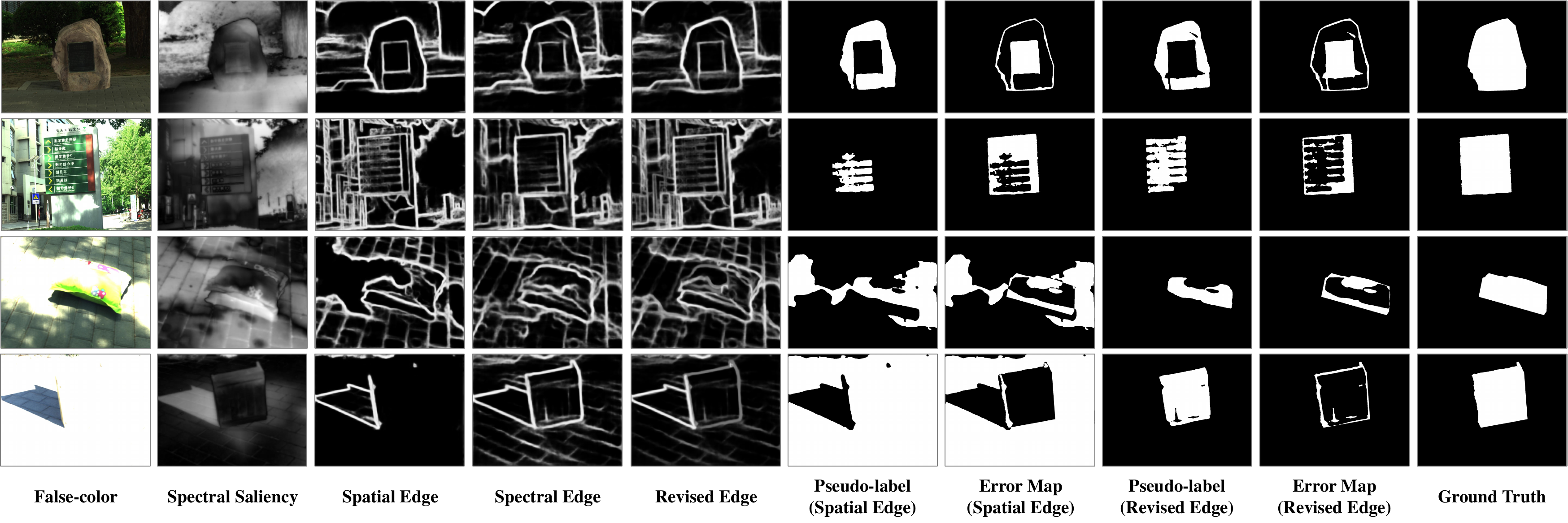}
    \caption{Illustration of edge map refinement and the generated pseudo-labels. Spectral Saliency can supplement false-color images' missing or discontinuous edge information, producing high accuracy and maximum coverage pseudo-labels.}
    \label{fig: edge and labels}
\end{figure*}

In the context of complex lighting conditions, RGB-image-based methods frequently encounter challenges, often either failing to detect the object entirely or producing imprecise outlines. While HSI-based methods are theoretically more resilient to lighting variations due to their reliance on spectral information, some still fall short in generating comprehensive and accurate detection in certain scenarios. This inconsistency is likely due to their limited capability in exploiting the abundant spectral information inherent in hyperspectral images. Our SPSD, in contrast, maximizes the use of spectrum through the entire process, thus demonstrating robust detection capabilities in such challenging scenarios. 

\begin{figure}[tp]
    \centering
    \includegraphics[width=\linewidth]{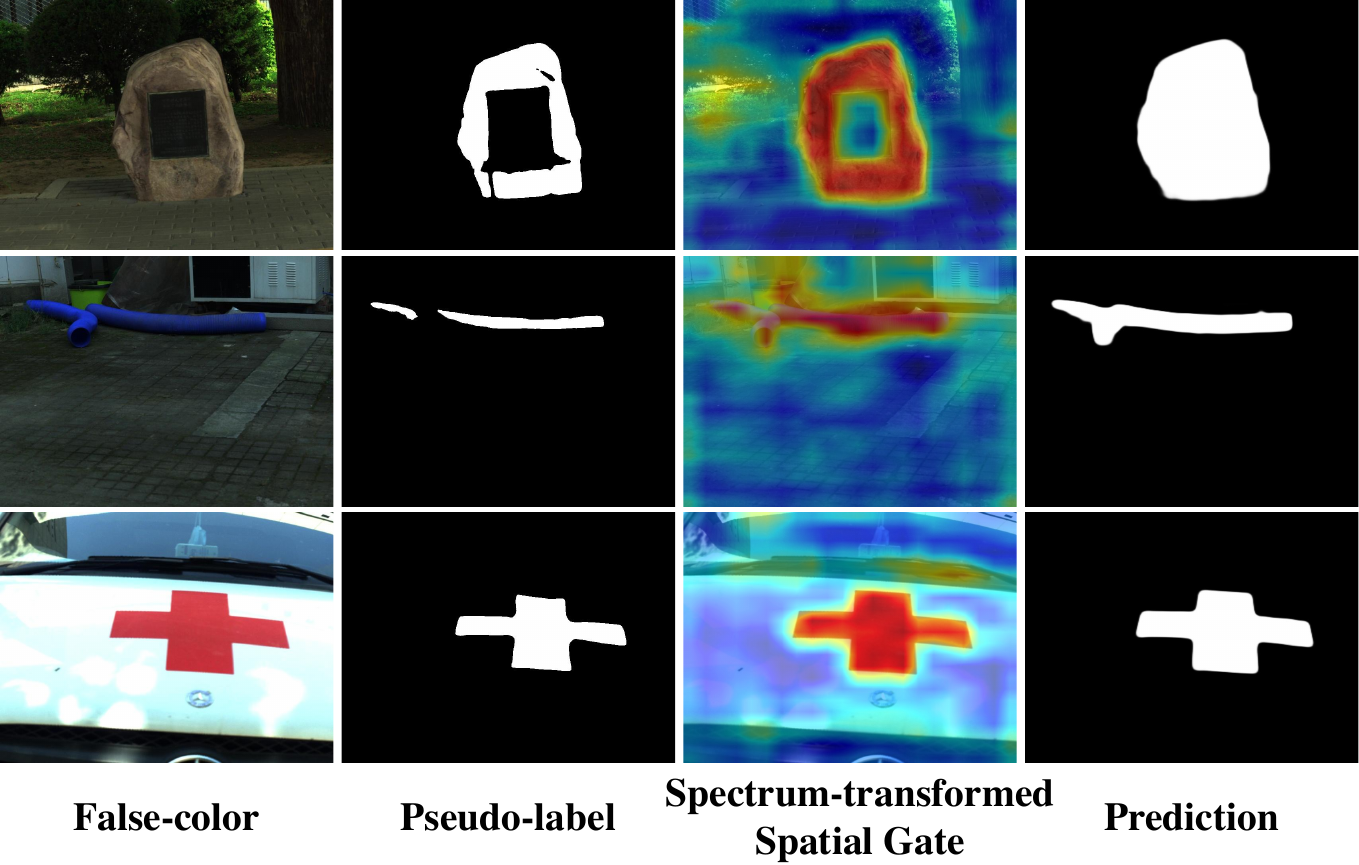}
    \vspace{-0.3cm}
    \caption{Visualization of the Spectrum-transformed Spatial Gate. It selectively attends to salient regions within the image, thereby augmenting the model's detection capability.}
    \vspace{-0.4cm}
    \label{fig: vis of gate}
\end{figure}

\nbf{Visualization of Pseudo-labels}
\cref{fig: edge and labels} showcases the enhancement of edge images through Spectral Saliency and the generated pseudo-labels. In the first row, the impact of Spectral Saliency is relatively minimal, as the false-color edge image alone is sufficient to isolate the target object. In contrast, the second row demonstrates a scenario where Spectral Saliency plays a crucial role in enhancing the completeness of object contours, which is essential for generating accurate and comprehensive pseudo-labels.

For scenes characterized by uneven illumination and overexposure, as depicted in the third and fourth rows, the role of Spectral Saliency becomes significantly more pronounced. In these cases, it supplements the missing edges of the false-color image, significantly improving the accuracy of the generated pseudo-labels. Comparison between error maps highlights the method's effectiveness in generating pseudo-labels.

\nbf{Visualization of Spectrum-transformed Spatial Gate}
The Spectrum-transformed Spatial Gate, denoted as $\boldsymbol{G}$, is visualized in \cref{fig: vis of gate}. It selectively amplifies saliency features within the model, thereby focusing more on salient regions. This process effectively reduces feature redundancy and enhances the model's detection capability.

\subsection{Results on HS-SOD}
\nbf{Quantitative Results} 
The quantitative results from our evaluation on the HS-SOD dataset are detailed in \cref{Dingliang}. Here, our SPSD method demonstrates superior performance over both traditional techniques and the SUDF method. It achieves an MAE of $0.073$, an $E_\xi$ of $0.778$, and an $F_\beta$ of $0.626$. Additionally, SPSD significantly outperforms the weakly supervised PSOD in other metrics except for AUC. This superior performance can be attributed to our method's ability to fully exploit the spectral information, resulting in enhanced feature extraction and more accurate saliency detection compared to traditional methods and SUDF.

The use of our pseudo-labels to train PSOD on the HS-SOD dataset led to a more substantial improvement in model detection performance compared to the enhancement observed on the HSOD-BIT dataset. Similarly, the degradation in detection performance when using PSOD' ’s pseudo-labels is less pronounced on the HS-SOD dataset than on the HSOD-BIT dataset. This is attributed to the HS-SOD dataset's larger proportion of sizable objects. In scenarios with such large objects, constraining the flood region to generate pseudo-labels can introduce considerable bias. This outcome strongly supports the necessity of employing spectral information to refine edges and avoiding restrictions on flood-filling regions. By integrating spectrum throughout the pseudo-label generation process, our method ensures that the labels are comprehensive and accurate, even for large objects, thereby enhancing the overall performance of the model.

\begin{figure*}
    \centering
    \includegraphics[width=\linewidth]{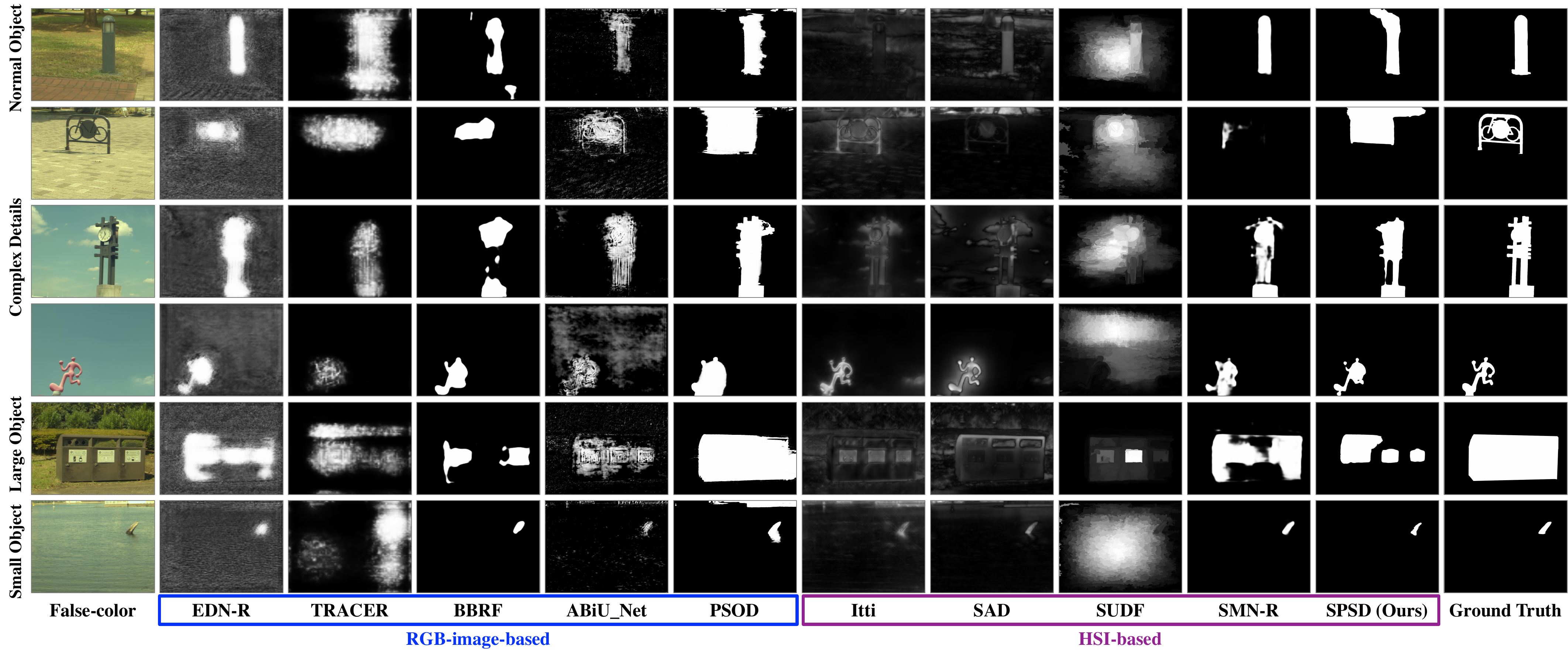}
    \caption{Qualitative results on HS-SOD. From top to bottom: normal object, objects with complex details, large object, and small object.}
    \label{fig: hssod-dingxing}
\end{figure*}

\begin{figure}[tp]
    \centering
    \includegraphics[width=1\linewidth]{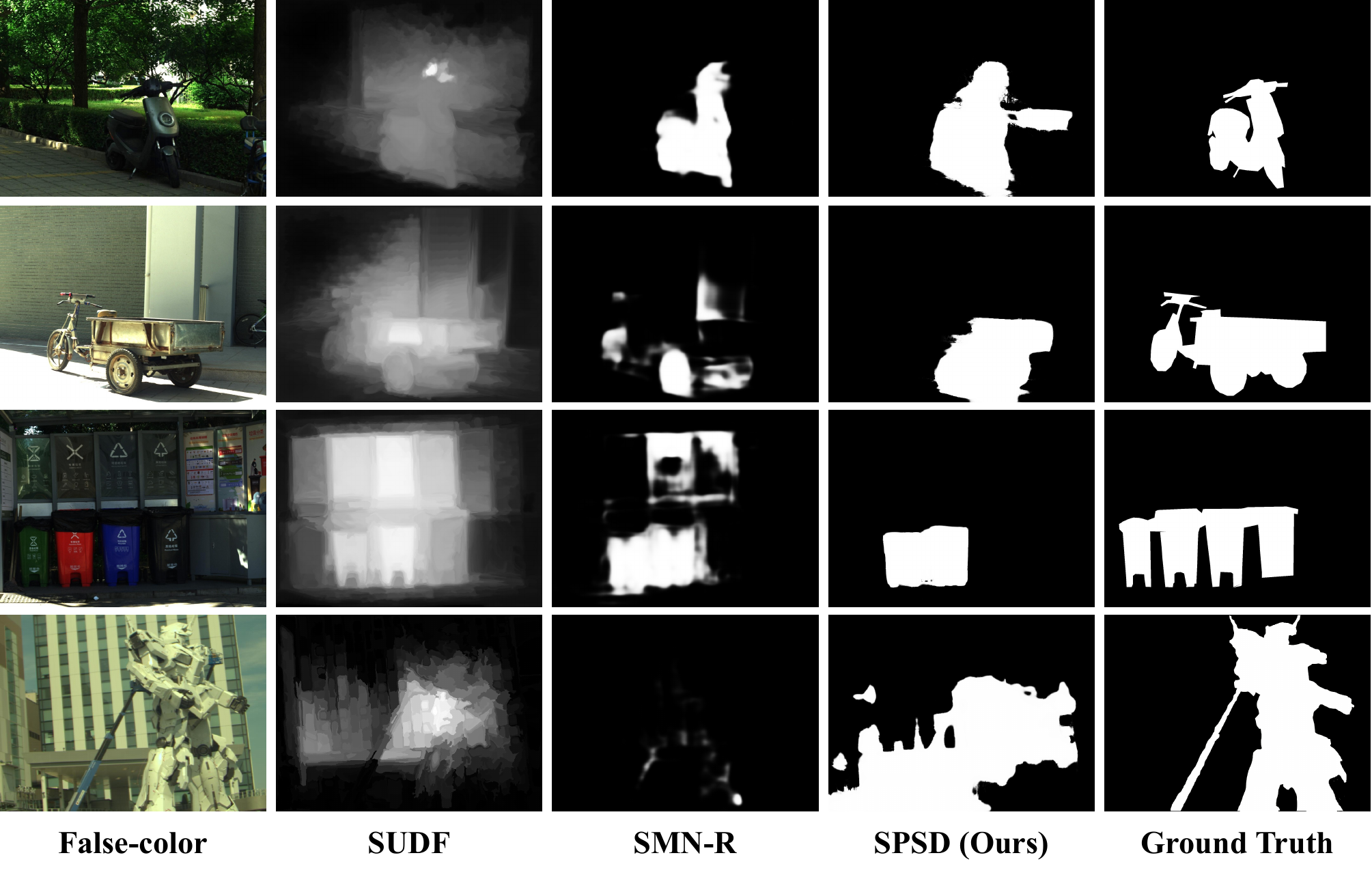}
    \caption{Illustration of failure cases. SPSD and other methods fail when targets have complex contours and hollow parts.}
    \label{fig:fail case}
\end{figure}

\nbf{Qualitative Results}
The qualitative results are shown in \cref{fig: hssod-dingxing}. Conventional methods and SUDF generally produce more erroneous and incomplete results. This is due to their reliance on hand-crafted features or shallow convolutional layers, which are often insufficient to capture the complex spectral information inherent in hyperspectral images.
When detecting objects with complex details, traditional algorithms manage to preserve these details by calculating spectral variations between layers. However, neural network-based methods struggle to accurately segregate salient objects from the background. This highlights the limitations of these neural networks in fully exploiting the spectrum, which is essential for accurate saliency detection in complex scenes. Our SPSD, by contrast, leverages the spectral information throughout the entire detection process, ensuring more accurate and comprehensive detection even in detailed scenes.
In other scenarios, SPSD lags slightly behind RGB-image-based methods, displaying mislabeled results and incompleteness. This slight decline can be attributed to the nature of point supervision, which provides less training data compared to fully supervised methods. However, the integration of spectral information in our approach still allows SPSD to achieve competitive results, demonstrating the robustness of our method even with limited supervision.

\subsection{Failure Cases}
Although SPSD surpasses existing HSOD methods and rarely produces errors, some failure cases remain, particularly with targets having overly complex contours and hollow parts, as shown in \cref{fig:fail case}. Specifically, in the first row, while SPSD correctly detects the bike, the contour delineation precision is low. In the last three rows, SPSD fails to detect the salient objects accurately. Similarly, other compared methods, such as SMN and SUDF, also fail in these scenarios. The presumed reasons are: (1) there methods are not designed for complex edges, and (2) such data is sufficient in the dataset.

\begin{figure*}[tp]
    \centering
    \includegraphics[width=\linewidth]{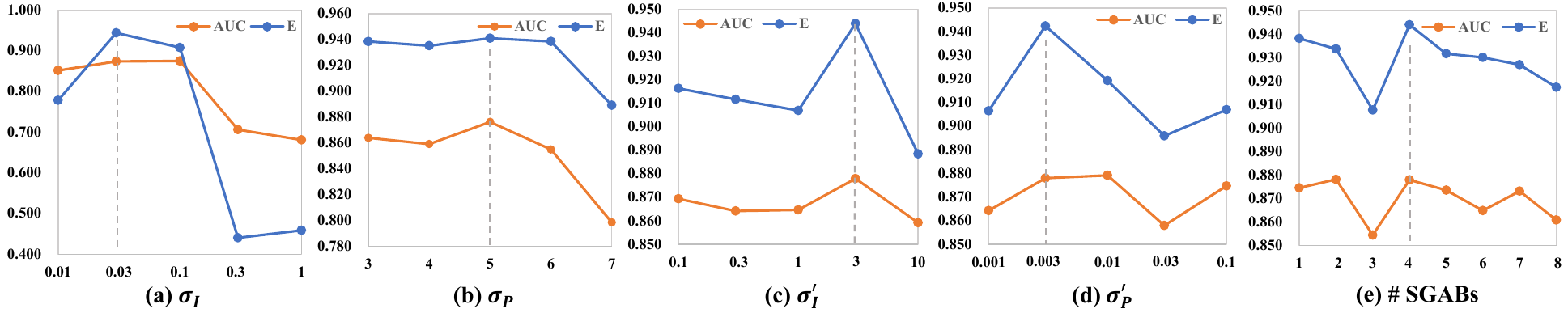}
    \caption{Parameter analysis of $\sigma_I$, $\sigma_P$, $\sigma_I^{\prime}$, $\sigma_P^{\prime}$, and the numbers of Saliency Guided Attention Blocks.}
    \label{fig: hyperparameter}
\end{figure*}

\begin{figure*}[tp]
    \centering
    \includegraphics[width=\linewidth]{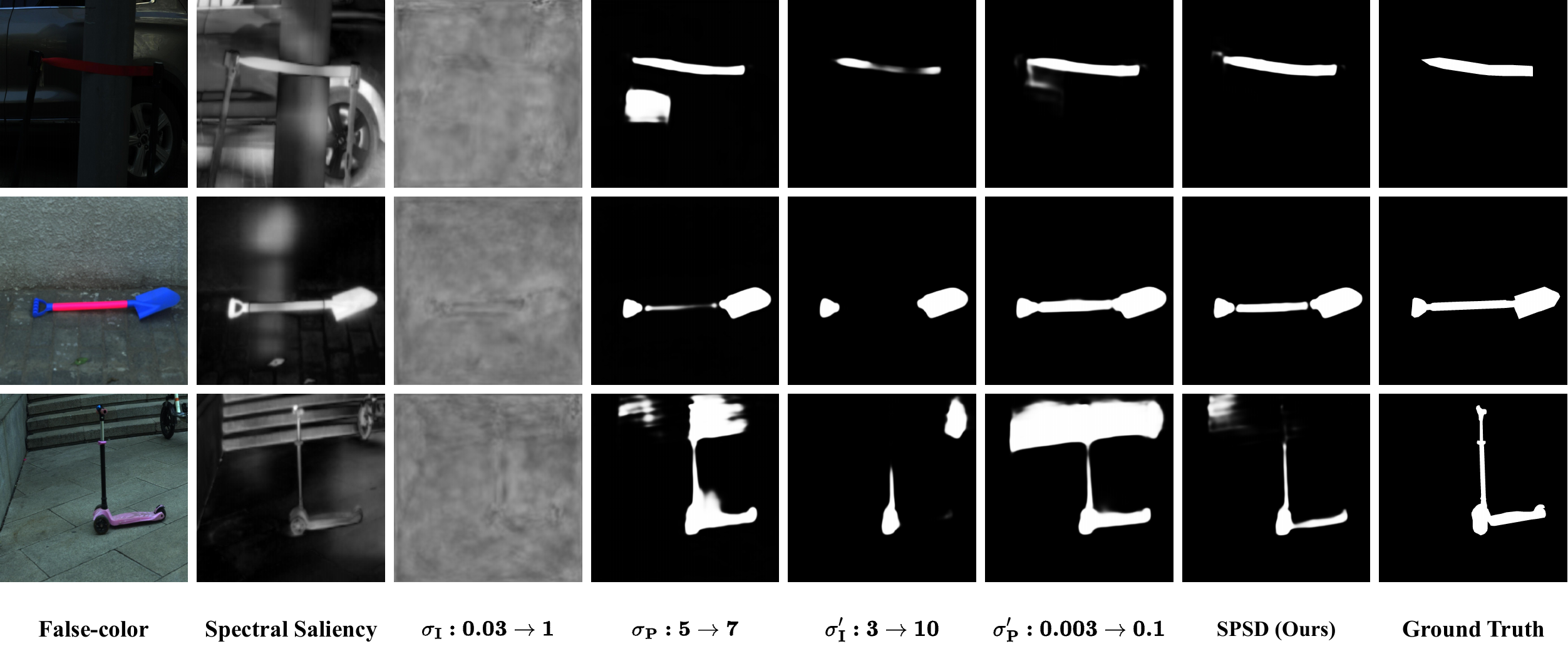}
    \caption{Visualization of the impact of $\sigma_I$, $\sigma_P$, $\sigma_I^{\prime}$, and  $\sigma_P^{\prime}$. Our SPSD adopts $\sigma_I = 0.03$, $\sigma_P = 5$, $\sigma_I^{\prime} = 3$, and $\sigma_P^{\prime} = 0.003$.
    }
    \label{fig: vis hyperparameter}
\end{figure*}

\subsection{Ablation Study}
\nbf{Hyperparameter Analysis}
Hyperparameters $\sigma_I$, $\sigma_P$, $\sigma_I^{\prime}$, and $\sigma_P^{\prime}$ can significantly influence the network's detection performance. Consequently, we conduct a detailed hyperparameter analysis, the results of which are presented in \cref{fig: hyperparameter}~(a)-(d). Specifically, we observe that with the increase in $\sigma_I$ and $\sigma_P$, the network's detection performance initially improves slightly, then deteriorates significantly, as measured by the AUC and $E_\xi$ metrics. Similarly, variations in $\sigma_I^{\prime}$ and $\sigma_P^{\prime}$ lead to fluctuations in performance, peaking at $\sigma_I^{\prime} = 3$ and $\sigma_P^{\prime} = 0.003$.
The effects of these hyperparameters are illustrated in \cref{fig: vis hyperparameter}. For example, with $\sigma_I = 1$, the model does not converge within 200 iterations, resulting in detection failure. At $\sigma_I^{\prime} = 10$, the model overemphasizes Spectral Saliency values for salient object detection, mistakenly identifying areas with similar Spectral Saliency responses as part of the foreground, thus impairing the accuracy of detection boundaries. Furthermore, when $\sigma_P = 7$ or $\sigma_P^{\prime} = 0.1$, the model tends to classify spatially adjacent objects as salient, causing a significant misclassification of background regions as foreground, compared to optimal settings (as depicted in the penultimate column).

Additionally, we conduct an ablation study examining the number of Saliency Guided Attention Blocks (\#SGABs), with findings presented in \cref{fig: hyperparameter}~(e). This study indicates a general decrease in detection performance as the number of SGABs increases, although some irregularities are observed. Optimal performance is achieved with a configuration of four SGABs.

Based on this comprehensive analysis, the ideal hyperparameter settings are determined to be $\sigma_I = 0.03$, $\sigma_P = 5$, $\sigma_I^{\prime} = 3$, $\sigma_P^{\prime} = 0.003$, with the number of SGABs fixed at 4.

\nbf{Impact of Scaling the Dimensions on Pseudo-labels}
Edge images derived from multi-scale representations exhibit unique traits: high-resolution images furnish edge maps replete with intricate details, whereas low-resolution counterparts better capture object contours. We subjected edge images of varying scales to the flood-fill algorithm and juxtaposed the resultant pseudo-labels against ground truth annotations. The quantitative assessment is tabulated in \cref{tab: multi-scale edge}. Notably, when image dimensions are scaled down to $0.50$ of their original size, the generated pseudo-labels closely approximate the annotated ground truth.

\begin{table}[]
    \centering
    \caption{Impact of Scaling the Dimensions on Pseudo-labels.}
    \setlength{\tabcolsep}{7mm}
    \begin{tabular}{c|ccc}
        \toprule[1.2pt]
        Scale  & MAE $\downarrow$   & $F_\beta$ $\uparrow$ & CC $\uparrow$  \\
        \midrule
        0.25 & \textbf{0.077} & 0.695    & 0.594 \\
        0.50 & 0.079 & \textbf{0.706}    & \textbf{0.603} \\
        0.75 & 0.085 & 0.676    & 0.582 \\
        1.00 & 0.091 & 0.659    & 0.565 \\
        1.25 & 0.110 & 0.640    & 0.547  \\
        \bottomrule[1.2pt]
    \end{tabular}
    \label{tab: multi-scale edge}
\end{table}

\nbf{Effect of Edge Refinement Using Spectral Saliency}
A comparative analysis between pseudo-labels generated solely from false-color images and those incorporating both false-color and Spectral Saliency images reveals the utility of Spectral Saliency. As delineated in \cref{tab: revision}, the edge refinement using Spectral Saliency leads to improvements across all evaluation metrics, thereby corroborating its efficacy. By integrating Spectral Saliency, our method effectively captures contours that are not apparent in false-color images alone, ensuring more precise edge delineation. This principle of combining false-color and spectrum enables the generation of higher-quality pseudo-labels, which significantly improves overall detection performance.

\begin{table}[tp]
    \centering
    \caption{Quantitative Results of Spectral Saliency refinement.}
    \setlength{\tabcolsep}{3.3mm}{
        \begin{tabular}{cc|ccc}
            \toprule[1.2pt]
            False-color & Spectral Saliency & MAE $\downarrow$  & $F_\beta$ $\uparrow$ & CC $\uparrow$  \\
            \midrule
            \color{green} \ding{51} & \color{red} \ding{55} & 0.091 & 0.696 & 0.602 \\
            \color{green} \ding{51} & \color{green} \ding{51} & \textbf{0.079} & \textbf{0.706} & \textbf{0.603} \\
            \bottomrule[1.2pt]
        \end{tabular}
    }
    \label{tab: revision}
\end{table}

\nbf{Effect of Hybrid CRF Loss}
The distinct contributions of false-color images and spectral saliency within the hybrid CRF loss are methodically evaluated, with the findings detailed in~\cref{tab: CRF loss ablation}. This loss function, which integrates spatial and intensity information from both false-color and spectral saliency images, plays a critical role in refining boundaries and enhancing spatial coherence in salient object detection results. This is especially important in point supervision settings, where precise boundary delineation is challenging. Consequently, omitting this loss leads to a significant reduction in detection performance, as illustrated in the first row of~\cref{tab: CRF loss ablation}. Additionally, the detailed color and texture information in false-color images provide a strong foundation for the CRF loss to operate on, enhancing the ability to detect and delineate objects accurately. Meanwhile, spectral saliency plays a supplementary role, contributing essential spectral information that enhances edge refinement and overall saliency accuracy. The complementary use of false-color and spectral saliency images within the CRF loss function highlights the effectiveness of our approach in leveraging the strengths of both data types to improve detection performance.

\begin{table}[tp]
    \centering
    \caption{The effectiveness of each component in hybrid CRF loss.}
    \label{tab: CRF loss ablation}
    \setlength{\tabcolsep}{3mm}{
        \begin{tabular}{cc|ccc}
            \toprule[1.2pt]
            False-color  & Spectral Saliency & MAE $\downarrow$   & $F_\beta$ $\uparrow$     & CC $\uparrow$    \\
            \midrule
            \color{red} \ding{55} & \color{red} \ding{55}  & 0.175 & 0.473 & 0.537 \\
            \color{green} \ding{51} & \color{red} \ding{55}  & 0.070 & 0.703 & 0.728 \\
            \color{red} \ding{55} & \color{green} \ding{51}  & 0.032 & 0.855 & 0.838 \\
            \color{green} \ding{51} & \color{green} \ding{51} & \textbf{0.031}  & \textbf{0.878} & \textbf{0.864}  \\
            \bottomrule[1.2pt]
        \end{tabular}
    }
\end{table}

\begin{table}[tp]
    \centering
    \caption{The effectiveness of each component of our model.}
    \label{tab: ablation}
    \setlength{\tabcolsep}{2.3mm}
        \begin{tabular}{cccc|ccc}
            \toprule[1.2pt]
            SRGM & GA & SGAB     & Sal. Ref.     & MAE $\downarrow$   & $F_\beta$ $\uparrow$     & CC $\uparrow$    \\
            \midrule
            \color{red} \ding{55} & \color{green} \ding{51} & \color{green} \ding{51} & \color{green} \ding{51}  & 0.032 & 0.851 & 0.850 \\
            \color{green} \ding{51} & \color{red} \ding{55} & \color{green} \ding{51} & \color{green} \ding{51}  & 0.034 & 0.846 & 0.817 \\
            \color{green} \ding{51} & \color{green} \ding{51} &  \color{red} \ding{55} & \color{green} \ding{51}  & 0.032 & 0.851 & 0.850 \\
            \color{green} \ding{51} & \color{green} \ding{51} & \color{green} \ding{51} & \color{red} \ding{55} & 0.033 & 0.856 & \textbf{0.869} \\
            \color{green} \ding{51} & \color{green} \ding{51} & \color{green} \ding{51} & \color{green} \ding{51}   & \textbf{0.031}  & \textbf{0.878} & 0.864  \\
            \bottomrule[1.2pt]
        \end{tabular}
\end{table}

\begin{table}[tp]
    \centering
    \caption{The impact of Spectral Saliency and the entire spectrum.}
    \label{tab: ablation2}
    \setlength{\tabcolsep}{2.5mm}{
        \begin{tabular}{ccc|ccc}
            \toprule[1.2pt]
            HSI & False-color & Spec. Sal. & MAE $\downarrow$  & $F_\beta$ $\uparrow$  & CC $\uparrow$ \\
            \midrule
            \color{green} \ding{51} & \color{red} \ding{55} & \color{red} \ding{55} & 0.049  & 0.752 & 0.723   \\
            \color{red} \ding{55} & \color{green} \ding{51} & \color{red} \ding{55} & 0.052 & 0.726 & 0.697   \\
            \color{red} \ding{55} & \color{green} \ding{51} & \color{green} \ding{51} & 0.040 & 0.785 & 0.781   \\
            \color{green} \ding{51} & \color{red} \ding{55} & \color{green} \ding{51} & \textbf{0.031}  & \textbf{0.878} & \textbf{0.864}  \\
            \bottomrule[1.2pt]
        \end{tabular}
    }
\end{table}

\nbf{Impact of Spatial Refinement and Gating Mechanism}
To assess the impact of SRGM, we conduct an experiment in which SRGM is removed, and the deep Spectral Saliency feature $\boldsymbol{F}_{S}$ is directly fed into the Saliency Guided Attention Block. This modification leads to the introduction of additional noise, resulting in a decline in detection performance. Specifically, as indicated in the first row of \cref{tab: ablation}, there is a notable decrease of $0.027$ in $F_\beta$, a reduction of $0.014$ in CC, and an increase of $0.001$ in MAE. These changes highlight the effectiveness that SRGM filters out noise and enhances the relevant spectral features, ensuring that only the most salient information is passed on to subsequent processing stages. This refinement is crucial for maintaining high detection accuracy, as well as reduce feature redundancy.
 
\nbf{Effect of Saliency Guided Attention Block}
We conduct two ablation studies to investigate the Guided Attention (GA) mechanism and SGAB, with results presented in the second and third rows of \cref{tab: ablation}, respectively. In the first scenario, we omit GA, using the Spectrum-transformed Spatial Gate $\boldsymbol{G}$ to directly multiply with the decoder's input. This alteration results in decreased scores in $F_\beta$ and CC, along with an increase in MAE, confirming the instrumental role of the Guided Attention mechanism in boosting detection accuracy. The Guided Attention mechanism ensures that the attention is focused on the most relevant features. Its absence leads to a less effective focus on these crucial features, thereby reducing the overall detection performance. In the second scenario, we completely remove the SGAB, employing $\boldsymbol{G}$ to modulate the deep HSI feature $\boldsymbol{F}_{I}$. By removing SGAB, the model loses its ability to effectively highlight these regions, resulting in diminished performance. This principle of using specialized attention mechanisms to refine and focus features is fundamental to our method, ensuring that the most relevant information is emphasized for superior detection accuracy.

\nbf{Effect of Saliency Refinement}
The impact of saliency refinement on the results is depicted in the fourth row of \cref{tab: ablation}. While omitting saliency refinement leads to a marginal increase in CC by 0.005, it simultaneously results in a marked reduction of the adaptive $F_\beta$ score by $0.022$. Saliency refinement ensures that the detected saliency maps are more precise and better aligned with the true object boundaries. The results clearly demonstrate that saliency refinement is a vital component in achieving high-quality detection results.

\nbf{Effect of Spectral Saliency}
To investigate the impact of Spectral Saliency images, we exclude them from SPSD, using only HSIs as input. Spectral Saliency images guide the model by enhancing the feature representation and ensuring that the attention mechanisms focus on the most salient regions. Without this guidance, the feature representation is less effective, leading to less accurate saliency detection. Besides, Spectral Saliency images provide additional information during the CRF refinement process. This supplementary spectral data helps refine the edges and improve the accuracy of the saliency maps. The absence of this information in the hybrid CRF loss restricts its effectiveness, resulting in poorer performance metrics. As demonstrated in the first row of \cref{tab: ablation2}, the absence of Spectral Saliency leads to a decrease in $F_\beta$ and CC, and an increase in MAE, thereby emphasizing the essential role of Spectral Saliency as an input.

\nbf{Effect of Spectrum}
To confirm the importance of the entire spectrum, we substitute the input HSI with the false-color image. Ensuring fairness, we randomly initialize the first layer of the backbone network. The results, detailed in the second row of \cref{tab: ablation2}, reveal that relying solely on false-color images leads to suboptimal detection performance. By incorporating Spectral Saliency, our method enhances the feature extraction process by leveraging the full spectral information, which significantly improves the model's ability to accurately detect and delineate salient objects. Spectral Saliency captures subtle spectral variations that are crucial for distinguishing salient regions from the background, thus providing a more comprehensive and precise input for the saliency detection model. This enhancement is evident in the results presented in the third row of \cref{tab: ablation2}. These experimental findings decisively highlight the effectiveness of utilizing the full spectrum in hyperspectral saliency detection. 

\subsection{Extended Experiment on RGB-thermal Datasets}
RGBT SOD is an emerging field that combines thermal imaging with RGB images to enhance the effectiveness of significant target detection. To validate the adaptability of our approach, we apply it to RGBT datasets and conduct extended experiments to evaluate its efficacy.

\nbf{Experimental Settings}
We make modifications to the initial layer of the backbone network $\boldsymbol{E}_{\theta}$ to accommodate 3-channel RGB images as input. Importantly, we abstain from using pre-trained weights for this layer to ensure a fair comparison. Additionally, we train SGSE under fully supervised condition to further assess its detection capability.

\nbf{Datasets} 
In line with Tu \etal~\cite{Tu_2022_RGBT}, SPSD is trained on VT5000's training set~\cite{Tu_2022_RGBT} and evaluated on VT821~\cite{Wang_2018_RGBT}, VT1000~\cite{Tu_2019_RGBT}, and VT5000's test set. We benchmark SPSD against MTMR~\cite{Wang_2018_RGBT}, EGNet~\cite{Zhao_2019_EGNet}, CPD~\cite{Wu_2019_Cascaded}, BASNet~\cite{Qin_2019_BASNet}, ADF~\cite{Tu_2022_RGBT}, MMNet~\cite{9439490}, and IFFNet~\cite{10015881}, employing maximum F-measure ($F_\beta$) and MAE as evaluation metrics.

\begin{table}[tp]
    \centering
    \caption{Results for RGBT SOD on VT821, VT1000 and VT5000 dataset.}
    \label{tab: RGBT exps}
    \setlength{\tabcolsep}{1.5mm}{
        \begin{tabular}{l|cc|cc|cc} 
            \toprule[1.2pt]
            \multirow{2}{*}{Method} & \multicolumn{2}{c|}{VT821} & \multicolumn{2}{c|}{VT1000} & \multicolumn{2}{c}{VT5000}  \\ 
            \cmidrule{2-7}
            & $F_\beta$ $\uparrow$ & MAE $\downarrow$ & $F_\beta$ $\uparrow$ & MAE $\downarrow$ & $F_\beta$ $\uparrow$ & MAE $\downarrow$ 
            \\ 
            \midrule
            MTMR~\cite{Wang_2018_RGBT}                    & 0.747 & 0.108              & 0.754 & 0.119               & 0.662 & 0.115               \\
            EGNet~\cite{Zhao_2019_EGNet}                   & 0.795 & 0.063              & 0.917 & 0.033               & 0.839 & 0.051               \\
            CPD~\cite{Wu_2019_Cascaded}                     & 0.786 & 0.079              & 0.914 & 0.031               & 0.847 & 0.047               \\
            BASNet~\cite{Qin_2019_BASNet}                  & 0.803 & 0.067              & 0.913 & 0.030               & 0.820 & 0.055               \\
            ADF~\cite{Tu_2022_RGBT}                     & 0.804 & 0.077              & 0.923 & 0.034               & 0.863 & 0.048               \\ 
            MMNet~\cite{9439490} & 0.867 & 0.040 & 0.920 & 0.027 & 0.852 & 0.042 \\
            IFFNet~\cite{10015881}  & \textbf{0.884} & \textbf{0.029} & \textbf{0.936} & \textbf{0.017} & \textbf{0.888} & \textbf{0.028} \\
            \midrule
            SPSD (Ours)             & 0.774 & 0.047              & 0.849 & 0.042               & 0.726 & 0.065               \\
            SGSE (Ours)                    & 0.826 & 0.040              & 0.910 & 0.025               & 0.836 & 0.041              \\
            \bottomrule[1.2pt]
        \end{tabular}
    }
\end{table}

\begin{figure}[tp]
    \centering
    \includegraphics[width=\linewidth]{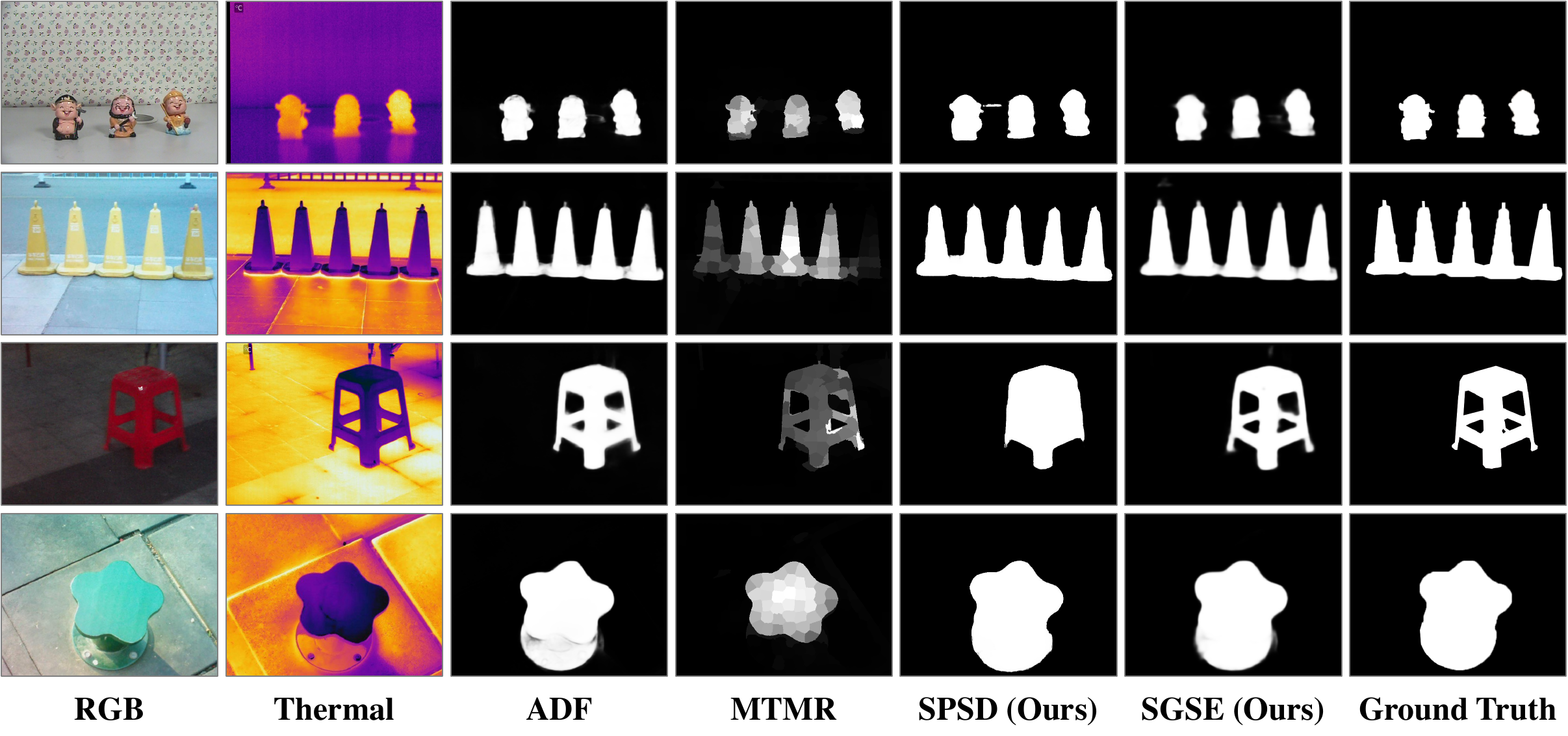}
    \caption{Qualitative results of RGBT SOD. Our method achieves reliable saliency detection results.}
    \label{fig: RGBT}
\end{figure}

\nbf{Quantitative Results}
Our SPSD achieves an $F_\beta$ score of $0.774$ on the VT821 dataset, slightly below that of EGNet. However, with an MAE of $0.047$, SPSD surpasses many fully supervised methods, ranking second only to MMNet. On the VT1000 and VT5000 datasets, SPSD slightly outperforms MTMR but falls short of EGNet. The reliance on single-point supervision in SPSD contributes to a noticeable performance gap compared to fully supervised methods.

Additionally, our SGSE demonstrates strong detection capabilities. On the VT821 dataset, SGSE outperforms ADF, attaining an $F_\beta$ of $0.825$ and an MAE of $0.040$. While SGSE’s $F_\beta$ score is higher than MTMR on both VT1000 and VT5000 datasets, its MAE closely approaches that of MMNet. These results validate the effectiveness and adaptability of our approach in RGBT salient object detection.

\nbf{Qualitative Results}
\cref{fig: RGBT} effectively illustrates the performance of our methods in RGBT SOD. SGSE achieves results comparable to ADF and MTMR, underscoring its competitive advantage. In contrast, SPSD, while showing potential, faces challenges in achieving detailed segmentation in the third scene. A side-by-side comparison with ADF and MTMR reveals that ADF misses segments in the fourth scene, whereas MTMR successfully detects salient objects.

\section{Future Work}
In future research, we aim to enhance the functionality and applicability of Spectrum-oriented Point-supervised Saliency Detector. One promising direction is the integration of depth information, which can significantly improve the robustness and accuracy of saliency detection and contour delineation. We will also focus on optimizing the computational efficiency of the integrated system to ensure real-time performance. This will involve developing advanced algorithms capable of concurrently processing hyperspectral and depth data without compromising speed. Additionally, we plan to explore practical applications of SPSD in various real-world scenarios, such as detecting and monitoring fire outbreaks in forests, assessing the growth status and health of crops, identifying camouflaged objects, and controlling the quality and safety of food products.

\section{Conclusion}
In this study, we propose the first point supervised method for saliency detection in hyperspectral images, namely Spectrum-oriented Point-supervised Saliency Detector. Motivated by introducing point supervision into hyperspectral salient object detection, we integrate Spectral Saliency into the framework for better hyperspectral image adaptation. Furthermore, we propose a Spectrum-transformed Spatial Gate to selectively enhance saliency features. Experimental results show that our method outperforms existing state-of-the-art HSOD methods and can rival fully supervised methods. Ablation experiments validate the effectiveness of each module, while RGBT SOD extension experiments confirm the generalization ability of our method.

\section{Acknowledgement}
This work was financially supported by the National Key Scientific Instrument and Equipment Development Project of China (No. 61527802), the National Natural Science Foundation of China (No. 62101032), the Young Elite Scientist Sponsorship Program of China Association for Science and Technology (No. YESS20220448), and the Young Elite Scientist Sponsorship Program of Beijing Association for Science and Technology (No. BYESS2022167).

\bibliographystyle{IEEEtran}
\bibliography{ref}
	
\end{document}